\documentclass[10pt,journal,compsoc]{IEEEtran}
\usepackage{graphicx}
\usepackage{epstopdf}
\usepackage{amsmath}
\usepackage{amssymb}
\usepackage{myalgorithm,myalgorithmic}
\usepackage{multirow}
\usepackage{hyperref}

\usepackage{graphicx}
\usepackage{epstopdf}
\usepackage{amsmath}
\usepackage{amssymb}
\usepackage{myalgorithm,myalgorithmic}
\usepackage{multirow}
\usepackage{makecell}
\usepackage{hyperref}
\usepackage{array}
\usepackage{booktabs}
\usepackage{color}

\ifCLASSINFOpdf
\else
\fi
\begin{document}
%
\title{Transfer Adaptation Learning: A Decade Survey}
%
%
%

\author{Lei Zhang,
        Xinbo Gao\\%

\IEEEcompsocitemizethanks{\IEEEcompsocthanksitem

L. Zhang is with the School of Microelectronics and Communication Engineering, Chongqing University, Chongqing 400044, China.
\ (E-mail: leizhang@cqu.edu.cn).
\IEEEcompsocthanksitem  X. Gao is with the Chongqing Key Laboratory of Image Cognition, Chongqing University of Posts and Telecommunications, Chongqing 400065.
 \ (E-mail: gaoxb@cqupt.edu.cn).

}}

%
%

\markboth{JOURNAL OF LATEX CLASS FILES,~Vol.~-, No.~-, MARCH~2019}%
{Shell \MakeLowercase{\textit{et al.}}: Bare Demo of IEEEtran.cls for IEEE Journals}
%




\IEEEtitleabstractindextext{
\begin{abstract}
The world we see is ever-changing and it always changes with people, things, and the environment. \textit{Domain} is referred to as the state of the world at a certain moment. A research problem is characterized as \textit{transfer adaptation learning (TAL)} when it needs knowledge correspondence between different moments/domains. Conventional machine learning aims to find a model with the minimum expected risk on test data by minimizing the regularized empirical risk on the training data, which, however, supposes that the training and test data share similar joint probability distribution. TAL aims to build models that can perform tasks of target domain by learning knowledge from a semantic related but distribution different source domain. It is an energetic research filed of increasing influence and importance, which is presenting a blowout publication trend. This paper surveys the advances of TAL methodologies in the past decade, and the technical challenges and essential problems of TAL have been observed and discussed with deep insights and new perspectives. Broader solutions of transfer adaptation learning being created by researchers are identified, i.e., instance re-weighting adaptation, feature adaptation, classifier adaptation, deep network adaptation and adversarial adaptation, which are beyond the early semi-supervised and unsupervised split. The survey helps researchers rapidly but comprehensively understand and identify the research foundation, research status, theoretical limitations, future challenges and under-studied issues (\textit{universality}, \textit{interpretability}, and \textit{credibility}) to be broken in the field toward universal representation and safe applications in open-world scenarios.
\end{abstract}
%
}

\maketitle

%
\IEEEdisplaynontitleabstractindextext

\section{Introduction}
%
%
%
%
\IEEEPARstart{V}{isual} understanding of an image or video is a longstanding and challenging problem in computer vision. Visual classification, as a fundamental problem of visual understanding, aims to recognize \textit{what} an image depicts. A solidified route of visual classification is to establish a learning model based on a well-labeled image dataset, which can be recognized as \textit{target} data (task). However, labeling a large number of target samples is cost-ineffective, because it consumes a lot of human resources and time expenses in labeling and becomes almost unrealistic for many domain specific tasks in real applications. Therefore, leveraging another sufficiently labeled, distribution different but semantic related \textit{source} domain for recognizing target samples is becoming an increasingly important topic.

With the explosive increase of multi-source data from Internet such as YouTube and Flickr, a large number of labeled web database can be easily crawled. It is thus natural to consider train a learning model using multi-source web data for recognizing target data. However, a prevailing problem is that distribution mismatch and domain shift~\cite{TorralbaEfros2011, Shimodaira2000Shift} across source and target domain often exist owing to various factors such as resolution, illumination, viewpoint, background, weather condition, etc. in computer vision. Therefore, the classification performance on the target task is dramatically degraded when the source data for training the classifier has different distribution from the target data. This is due to that the fundamental independent identical distribution (i.i.d.) condition implied in statistical machine learning is no longer satisfied, which, therefore promotes the emergence of transfer learning (TL) and domain adaptation (DA)~\cite{YangYanHauptmann2007, PanYang2010,Saenko2010symm}. Originally, the concept of \textit{transfer} was first proposed in 1991 by Pratt et al.~\cite{Pratt1991aaai} and further introduced in~\cite{Pratt1993nips} for transfer between neural networks. Mathematically, the earlier TL supposed different joint probability distribution, i.e., $P(\mathcal{X}_{source}, \mathcal{Y}_{source}) \neq P(\mathcal{X}_{target}, \mathcal{Y}_{target})$ between source and target domains. DA supposed different marginal distribution, i.e., $P(\mathcal{X}_{source}) \neq P(\mathcal{X}_{target})$ but similar category space between domains i.e., $P(\mathcal{Y}_{source}|\mathcal{X}_{source})=P(\mathcal{Y}_{target}|\mathcal{X}_{target})$. Currently, according to different settings of label set between domains, a preliminary split of semi-supervised domain adaptation (SSDA), unsupervised domain adaptation (UDA), open-set domain adaptation (OSDA) and partial domain adaptation (PDA) is presented, in all which the source labels are completely available. Notably, both SSDA and UDA assume the \textit{same} category space $\mathcal{C}$ across domain (close-set), i.e., $\mathcal{C}_S=\mathcal{C}_T$. The former assumes a few labeled target samples can be used, while the target labels are completely unknown for the latter. Contrastively, OSDA and PDA assume the \textit{different} category space across domains, i.e. $\mathcal{C}_S\neq\mathcal{C}_T$. In general, OSDA supposes $\mathcal{C}_S\subset\mathcal{C}_T$ and PDA supposes $\mathcal{C}_T\subset\mathcal{C}_S$. In this survey, we aim to cover the recent progress from a pure methodological perspective beyond the conventional split of semi-supervised, unsupervised, open-set and partial DA tasks.

\begin{table}[htbp]
\centering
\caption{Summarization of Related Surveys in TL/DA}
\begin{tabular}{|c|c|c|c|c|c|}
 \hline
 No. & Survey Title & Ref. & Year \\
 \hline
 1 & \thead{A Survey on Transfer Learning} & \cite{PanYang2010} & 2010 \\
 \hline
 2 & \thead{Extreme learning machine based transfer \\learning algorithms:A survey} & \cite{Syed2017TL} & 2017 \\
 \hline
 3 & \thead{A survey of transfer learning} & \cite{Karl2016TL} & 2016 \\
 \hline
 4 & \thead{Transfer Learning for Visual\\ Categorization: A Survey} & \cite{Shao2015TL} & 2015 \\
 \hline
 5 & \thead{A survey of transfer learning for collaborative\\ recommendation with auxiliary data} & \cite{Pan2016TL} & 2016 \\
 \hline
 6 & \thead{Visual Domain Adaptation \\A survey of recent advances} & \cite{Patel2015DL} & 2015 \\
 \hline
 7 & \thead{Transfer learning using computational\\ intelligence: A survey} & \cite{Lu2015TL} & 2015 \\
 \hline
 8 & \thead{Deep visual domain adaptation: A survey} & \cite{Wang2018DA} & 2015 \\
 \hline
 9 & \thead{Transfer learning for activity\\ recognition: a survey} & \cite{Cook2013TL} & 2013 \\
\hline
10 & \thead{Domain adaptation for visual \\ applications: A comprehensive survey} & \cite{Csurka2017aXiv} & 2017 \\
\hline
\end{tabular}
\label{tab_review}
\end{table}

\begin{figure}
\centering
\includegraphics[height=2.3cm]{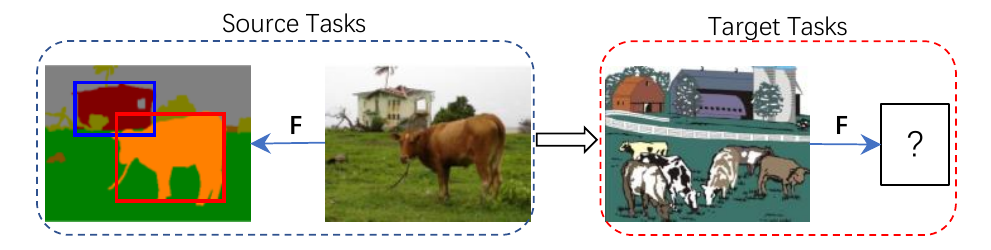}
\caption{Cross-domain object detection, recognition and semantic segmentation. $\mathbf{F}$ denotes models of three tasks learned on source domain. The value of TAL depends on the performance gain of target tasks.}
\label{fig1}
\end{figure}

The earlier related reviews on transfer learning and domain adaptation can be referred to as~\cite{PanYang2010, Cook2013TL,Lu2015TL,Patel2015DL,Shao2015TL, Pan2016TL,Wang2018DA,Karl2016TL,Syed2017TL,Csurka2017aXiv}, which are summarized in Table~\ref{tab_review}. However, our concern is that, these existing surveys can only reflect the several aspects in TL/DA community. With the recent explosive increase of transfer learning and domain adaptation models and algorithms, these surveys are unable to accurately characterize her status, challenges and future for this community from a more comprehensive perspective. Therefore, in this survey, we use a general name \textit{Transfer Adaptation Learning (TAL)} for unifying both TLs and DAs from a broader perspective, and discuss the technical advances, challenges and under-studied issues. In the past decade, especially the recent five years, TAL was one of the most active areas in machine learning community, and the goal of which is to narrow down the marginal and conditional distribution gap between source and target data, such that the labeled source data from one or more relevant domains can be utilized for executing different learning tasks in target domain, as illustrated in Fig.~\ref{fig1}. The cross-domain recognition, detection and segmentation tasks and challenges are highlighted.

%

Moving forward, deep learning (DL) techniques~\cite{Hinton2006Net, LeCun2015DL, Alex2012Net, He2015Resnet} have recently become dominant and powerful algorithms in feature representation and abstraction for image classification. In particular, the parameter adaptability and generality of DL models to other target data is worthy of praise, by fine-tuning a pre-trained deep neural network using a small amount of labeled target data. Therefore, \textit{fine-tune} has become a commonly used strategy for training deep models and frameworks in various applications, such as object detection~\cite{Ren2015Faster, Dai2016RFCN, Liu2016SSD, Fu2017DSSD}, person re-identification~\cite{Ahmed2015DL, Chung2017Siamese, Cheng2016Trip}, medical imaging~\cite{Hou2016Med, Esteva2017DNN, Shen2017Med}, remote sensing~\cite{Xie2016Poverty, Jean2016Poverty, Chen2016Hyper, Maggiori2017Conv}, etc. Generally, the \textit{fine-tune} can be recognized as a prototype for bridging the sufficiently labeled big source data and the few/un-labeled target data~\cite{Bengio2012JMLR}, which also facilitates the research progress of visual transfer adaptation learning (VTAL) in computer vision. Conceptually, \textit{fine-tune} is an intuitive and data-driven transfer learning method, which depends on pre-trained models with task-specific source database (e.g., ImageNet). The context of transfer learning challenge and why the generated representation with pre-trained models is useful have been formally explored in~\cite{Bengio2012JMLR}. Extensively, from the viewpoint of \textit{generative} learning, the popular generative adversarial net (GAN)~\cite{Goodfellow2014GAN} and its variants~\cite{Mirza2014CGAN,TranL2018DRGAN, Tran2017DRGAN, Reed2016GAN, Radford2016GAN} that aim to synthesize plausible images belonging to target distribution from some source distribution (e.g., noise signal), can also be recognized as generalized transfer learning techniques (e.g., style transfer tasks). Differently, conventional transfer learning approaches put emphasis on the \textit{output} knowledge adaptation (e.g., high-level features) across source and target domains, while GANs focus on \textit{input} data adaptation (e.g., low-level image pixels) from source distribution to target distribution. Recently, image pixel-level transfer has been intensively studied in image-to-image translation~\cite{Isola2017Trans, Zhu2017Unpaired, Choi2018Star, Liu2017Trans, Yoo2016Pixel}, style transfer~\cite{Gatys2016Style, Gatys2017Style, Johnson2016Style} and target face synthesis (e.g., pose transfer vs. age transfer)~\cite{Hu2018pose, Yin2017Towards, Cao2018Load, Huang2017Beyond, Yang2018Age}, etc.

In this survey, we focus on elaborating the technical advances and remained challenges in model-driven transfer adaptation learning. Learning from multiple sources for transferring or adapting to new target domains offers the possibility of promoting model generalization and understanding the biological learning essence. Notably, transfer adaptation learning is similar but different from multi-task learning that resorts to the shared feature representations or classifiers for multiple different tasks (e.g., detection vs. retrieval)~\cite{Evgeniou2007mtl,ZhangL2020TPAMI}, simultaneously.

In the past decade, a number of transfer learning and domain adaptation approaches have been emerged by following the expected target error upper-bounds of domain adaptation that minimizes a convex combination of the source empirical risk, domain discrepancy and joint error have been analyzed and justified~\cite{BenDavid2007NIPS,Blitzer2008NIPS}. Specifically, given a labeled source domain $\mathcal{D}_s$, an unlabeled target domain $\mathcal{D}_t$ and a hypothesis $h\in \mathcal{H}$ (hypothesis set), according to the Ben-David theorem~\cite{BenDavid2007NIPS}, the expected target risk $\epsilon_T(h)$ can be upper-bounded by the expected source risk $\epsilon_S(h)$, domain discrepancy $d_{\mathcal{H}\Delta\mathcal{H}}(\mathcal{D}_s, \mathcal{D}_t)$ and joint error $\lambda=\min_{h\in \mathcal{H}} \epsilon_S(h)+\epsilon_T(h)$. Intuitively, domain adaptation can be achieved by modeling and simultaneously minimizing the upper-bound of $\epsilon_T(h)$ from multiple different aspects, such as source error $\epsilon_S$, domain discrepancy $d(\mathcal{D}_s,\mathcal{D}_t)$ and joint error $\lambda$ under an ideal hypothesis $h^*$.

Due to the endless emergence and variety of existing methods of visual transfer adaptation learning (VTAL), in this paper, the remained challenges and main-stream methodological advances in VTAL community are identified and surveyed without further discussing the domain adaptation theory. Specifically, around the ultimate challenge of universal representation, considering the covariate shift, domain discrepancy and class discrimination, we propose five main-stream key technical categories of transfer adaptation learning for domain adaptive feature representations and classifiers, which are beyond the early split of semi-supervised and unsupervised (close-set) and recent split of open-set and partial domain adaptation. The novel taxonomy of VTAL is summarized as follows.
\begin{itemize}
\item \textit{Instance Re-weighting Adaptation}. Due to the probability distribution discrepancy across domains, it is natural to account for the difference by directly inferring the resampling weights of instances based on feature distribution matching in a non-parametric manner. Probability and Shannon entropy based re-weighting has become a common technique in UDA, OSDA and PDA problems. The \textit{adaptive} parameter estimation of the weights under a parametric distribution assumption remains to be a challenge.
\item \textit{Feature Adaptation}. For adapting the data from multiple sources, learning a common feature subspace or representation where the projected source and target domain are with similar distribution is generally resulted. However, learning common subspace generally belongs to shallow methodology, which relies on pre-trained deep models for discriminative feature representation. It is challenging to unify the shallow model with the deep model. Also, the heterogeneity of data distribution makes it challenging to gain such generic feature.
\item \textit{Classifier Adaptation}. The classifier trained on instances of source domain is often biased when recognizing instances from target domain due to the domain shifts and category space inconsistency. Learning a generalized classifier from source domain that can be used for recognizing \textit{seen/partially seen/unseen} object classes in another domain without bias is an essential but challenging topic.
\item \textit{Deep Network Adaptation}. Deep neural networks have been recognized with strong feature representation power and general deep model is built on single domain. Large domain shift makes deep neural network training challenging to obtain transferrable deep representation. Additionally, the \textit{unavailability (uncertainty)} of target labels (pseudo-labels) makes deep network adaptation still challenging due to negative transfer.
\item \textit{Adversarial Adaptation}. Adversarial learning originates from the generative adversarial nets, which has been a general technique in VTAL. A basic objective of VTAL is to make the source and target domains more close in marginal distribution. Therefore, it is amount to confusing the two domains, such that they can not be easily discriminated. However, due to the \textit{class mis-alignment} problem, there comes a technical challenge in class conditional domain confusion by using adversarial training and gaming strategy.
\end{itemize}

For each challenge, the taxonomic classes and sub-classes are presented to structure the recent work in transfer adaptation learning. We start with a discussion of weakly-supervised learning perspectives in Section 2, which is followed by the technical advances of TAL, including instance re-weighting adaptation (Section 3), feature adaptation (Section 4), classifier adaptation (Section 5), deep network adaptation (Section 6), and adversarial adaptation (Section 7). The existing benchmarks and future challenging tasks are discussed in Section 8. Extension of the versatile TAL theory and algorithm in more vision tasks is presented in Section 9. A conclusion of the research status and an outlook of the field for further promoting the healthy development of TAL are provided in Section 10.

\section{Weakly-supervised Learning Perspective}

Consider that VTAL can be characterised as weak-supervised learning, in this section, we first present some weakly-supervised learning perspectives, including semi-supervised, active learning, zero/few-shot learning and open set recognition. These learning methodologies are similar but essentially different from TAL, which, therefore, are briefly introduced for having a taste in this survey. Note that, in this section, unsupervised/self-supervised learning is not discussed, considering that the category label is completely unavailable which has different starting point from TAL. It must be mentioned that TAL does share essentially the same goal as unsupervise/self-supervised learning, i.e. learning generalized/universal representation.

The concept of \textit{weak learning} originated 20 years ago in AdaBoost~\cite{Ada1997Boost} and Ensemble learning~\cite{Ensemble1997Learn} algorithms, which tend to ensemble multiple weak learners to solve a problem. AdaBoost, that has been listed as the top 10 algorithms in data mining~\cite{Wu2008Top}, aims to learn multiple weak learners, in which each weak learner is obtained by training on the weighted incorrectly classified examples. By ensemble of multiple weak learners, the performance is significantly boosted. Although the \textit{weak} concept was proposed as early as 1997, the problem in that era was still established on strong supervision due to the relatively \textit{smaller} data. That is, the early problem can be strongly learned by conventional statistical learning models. However, today, the big data era, the problem becomes really a weak supervision problem, due to the \textit{inaccurate}, \textit{inexact}, and \textit{incomplete} characteristics of data labels~\cite{Zhou2017Weak}, which, therefore, has to be weakly learned. Currently, weakly-supervised learning is becoming a leading research topic. Undoubtedly, transfer adaptation learning, that resorts to solving cross-domain problems, is also a kind of weakly-supervised learning methodology. This section is deployed with typical weakly-supervised learning frameworks and perspectives.

\subsection{Semi-supervised Learning}

Semi-supervised learning (SSL) aims to solve the problem where there are a large amount of unlabeled examples $X_u$ and a few labeled examples $(X_l, Y_l)$ in the dataset~\cite{Chapelle2006SSL, Zhou2010SSL}. Generally, semi-supervised learning methods consist of four categories. (i) Generative methods that advocate generating the labeled and unlabeled data via an inherent model~\cite{Nigam2000SSL, Miller1997SSL, Kingma2014SSL}. (ii) Low-density separation methods that constrains the classifier boundary crossing the low-density region~\cite{Chapelle2005SSL, Joachims1999SSL, Li2013SSL}. (iii) Disagreement based methods that advocate co-training of multiple learners for annotating the unlabeled instances~\cite{Zhou2005Tri, Blum2005Co, Zhou2008Dis}. (iv) Graph based methods that propose to build the connection graph of the training instances for label propagation through graph modeling~\cite{Blum2001Graph, Belkin2004SSL, Kipf2017SSL, Yang2016SSL}. A good literature review of semi-supervised learning can be referred to as~\cite{Zhu2008SSL, Triguero2015SSL}.

Consider a general SSL framework, then the following expected risk is generally minimized.
\begin{equation}
\begin{split}
 &R\lbrack P_r, W, l(X, Y, W) \rbrack={E}_{(X,Y)\thicksim P_r} \lbrack l(X, Y, W)\rbrack\\
\end{split}
\end{equation}
where $P_r$ is the probability distribution, $X=\lbrack X_l, X_u \rbrack \in \Re^{D\times N}$ is the data, $Y=\lbrack Y_l, 0 \rbrack \in \Re^{C\times N}$ is the label index in which zero vector is posed for unlabeled samples, $W$ is the model parameter. $D, N$ and $C$ denote the number of dimensionality, samples, and classes of data, respectively.

The training data usually comes from a subset, therefore, the regularized risk, i.e., the average empirical risk with regularization is minimized.
\begin{equation}
\begin{split}
 &R_{reg}\lbrack W, l(X, Y, W) \rbrack=R_{emp}\lbrack W, l(X, Y, W) \rbrack + \lambda \Omega(W)\\
\end{split}
\end{equation}
where $l(\cdot)$ is the prediction loss function and $R_{emp}\lbrack \cdot \rbrack$ is the average empirical risk (e.g. mean squared loss) on training data.
A general SSL model with graph based manifold regularization can be written as
\begin{equation}
\begin{split}
 &\min_W R_{reg}\lbrack W, l(X, Y, W) \rbrack +\gamma \sum_{i, j}^{N}{A_{i,j}}d^2(f_i,f_j)\\
\end{split}
\end{equation}
where $f_i$ is the predicted label for sample $i$ and $A$ is the affinity matrix used for locality preservation. Usually, $A_{i,j}=\exp(-\sigma d^2(x_i,x_j))$ if $x_i$ and $x_j$ are neighbors, otherwise 0.

\begin{figure}
\centering
\includegraphics[height=2.5cm]{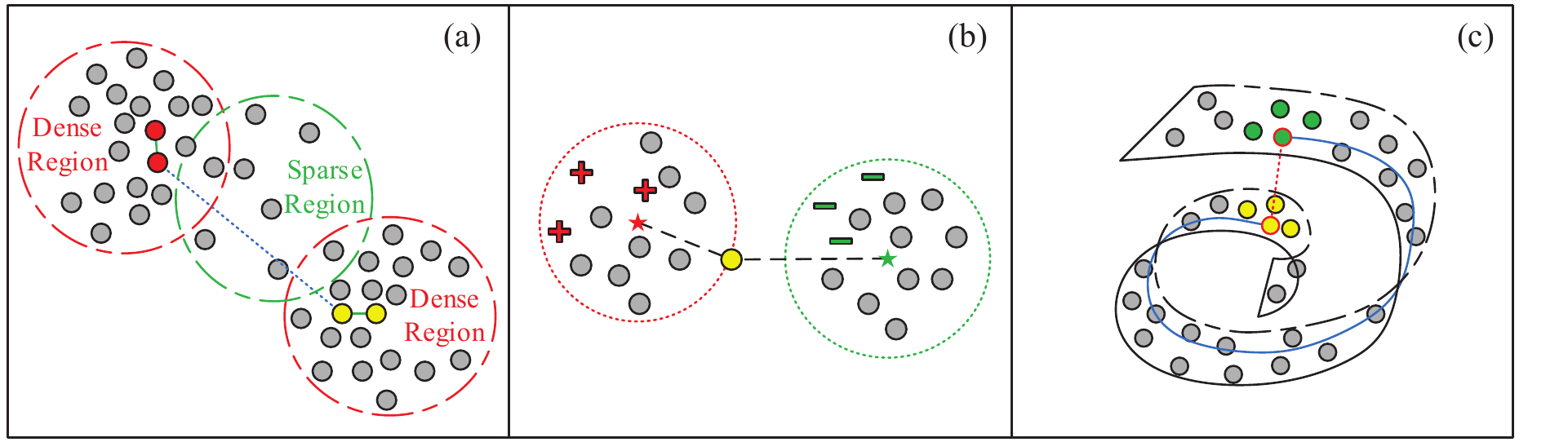}
\caption{Illustration of the three basic assumptions in SSL. a) \textit{Smoothness} assumption. b) \textit{Cluster} assumption. c) \textit{Manifold} assumption.}
\label{fig2}
\end{figure}

The key difference from transfer learning is that the marginal distribution and label space distribution are the same, i.e., $P(X_l)=P(X_u)$ and $P(Y_l|X_l)=P(Y_u|X_u)$. Generally, SSL attempts to exploit the unlabeled data for auxiliary training on the labeled data without human intervention, because the distribution of unlabeled data can intrinsically reflect sample class information. Actually, in SSL model, three basic assumptions, i.e., \textit{smoothness}, \textit{cluster}, and \textit{manifold}, have been established. The \textit{smoothness} assumption denotes that data is distributed with different density, and the two instances falling into the region of high density have the same label. The \textit{cluster} assumption denotes that data have inherent cluster structure, and the two samples in the same cluster are more similar. The \textit{manifold} assumption means that the data lie on a manifold, and the instances in a small local neighborhood have similar semantics. The three basic assumptions are visually shown in Fig.~\ref{fig2}.

\subsection{Active Learning}

Active learning (AL) aims to obtain the ground-truth labels of selected unlabeled instances with human intervention~\cite{Lewis1994A, Tong2000Support}, which is different from semi-supervised learning that exploits unlabeled data together with labeled data for improving recognition performance. Specifically, AL aims at progressively selecting and annotating the most informative data points from the pool of unlabeled samples, such that the labeling cost for training an effective model can be minimized~\cite{Elhamifar2013A, Settle2010Active}. There are two engines, learning engine and selection engine, in active learning paradigm. The learning engine targets at obtaining a baseline classifier, while the selection engine tries to select unlabeled instances and deliver them to human experts for manual annotation. The selection criteria is generally determined based on information uncertainty~\cite{Lewis1994A, Tong2000Support}.

\subsection{Zero/Few-shot Learning}

Recently, zero/few-shot learning (Z/FSL)~\cite{Rohrbach2011ZSL, Lampert2014ZSL, Fu2015ZSL, Ding2017ZSL,Niu2018ZSL}, as a typical weakly-supervised learning paradigm, has attracted researchers' attention. ZSL tries to recognize the samples of unseen categories that never appear in training data, i.e., there is no overlap between the seen categories in training data and the unseen categories in test data. That is, the label space distribution between training and test data is different, i.e., $P(Y_{seen}|X_{seen})\neq P(Y_{unseen}|X_{unseen})$, which can be recognized as a special case of transfer learning. This situation often occurs in various fields, due to that manually annotating tens of thousands of different object classes in the world is quite expensive and almost unrealistic. The general problem of ZSL is as follows.

\textbf{Zero-shot learning with disjoint training and testing classes}. \textit{Let $\mathcal{X}$ be an arbitrary feature space of training data. Let $\mathcal{Y}$ and $\mathcal{Z}$ be the sets of seen and unseen object categories, respectively, and there is $\mathcal{Y} \bigcap \mathcal{Z}=\emptyset$. The task is to learn a classifier f: $\mathcal{X}\mapsto\mathcal{Z}$ by using the training data $(x_1, y_1), \cdots, (x_N, y_N) \subset (\mathcal{X}, \mathcal{Y})$}.



An extension of ZSL is the one/few shot learning (O/FSL) where few labeled examples of each unseen object classes are revealed during training process. The usual idea of Z/O/FSL is to learn the embedding of the image feature into the semantic space or semantic attributes~\cite{Lampert2014ZSL, Rahman2018ZSL}. Afterwards, recognition of new classes can be conducted by matching the semantic embedding of the visual features with the semantic/attribute representation. However, visual-semantic mapping learned from the seen categories may not generalize well to the unseen category due to the domain shift, which, thus can be a challenging topic by utilizing transfer learning to ZSL. Actually, for improving ZSL under domain shifts, transductive or semi-supervised zero-shot learning approaches have been studied for reducing the difference of visual-semantic mappings between seen and unseen categories~\cite{Yu2018TZSL1, Yu2018TZSL2, Xu2018TZSL, Li2015SZSL, Song2018Trans}.

\begin{figure*}
\centering
\includegraphics[height=16cm]{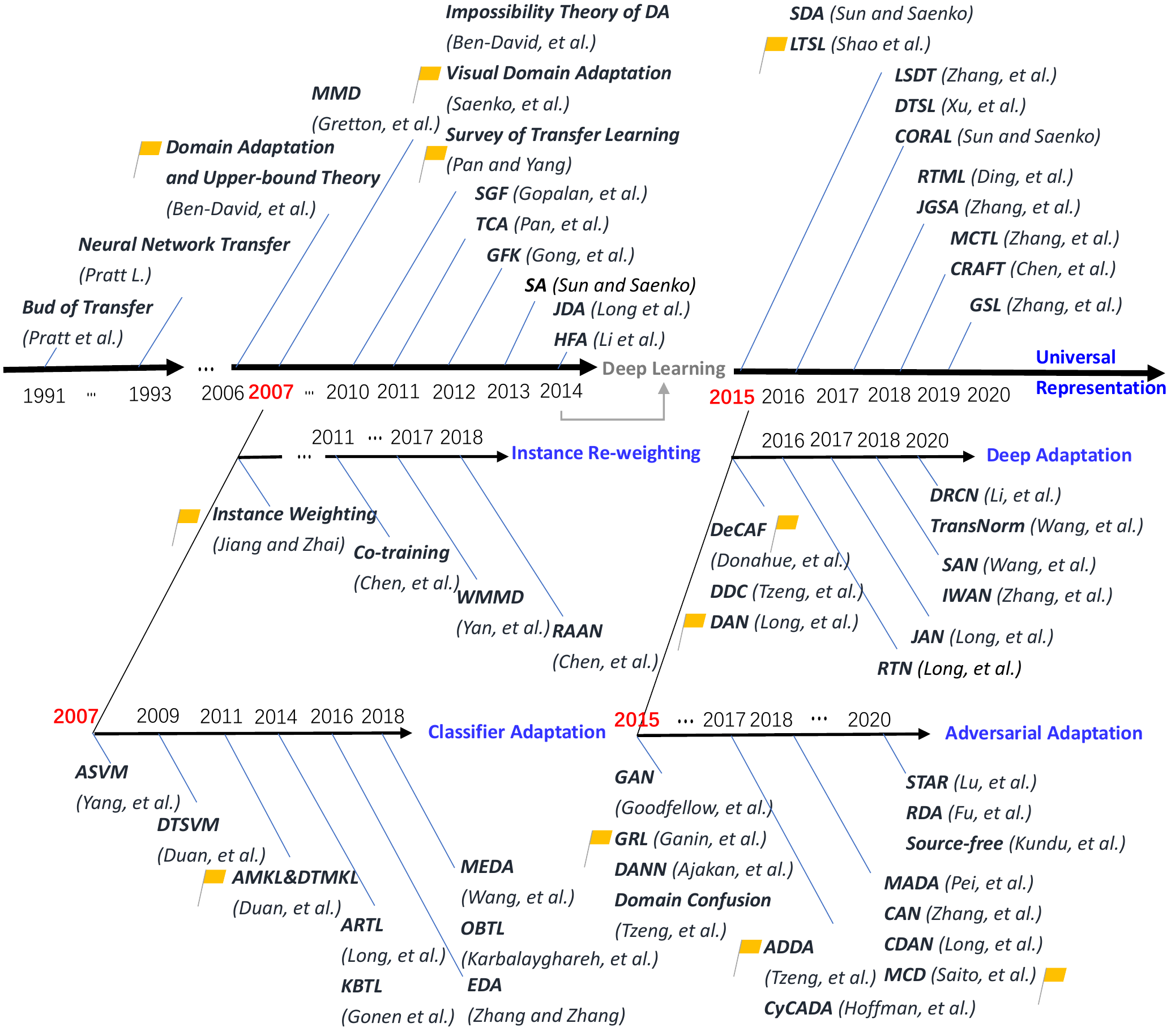}
\caption{A road map and research context of transfer adaptation learning. We represent the feature adaptation as the main stream which is divided into three stages (1991$\sim$1993, 2006$\sim$2015, 2015$\sim$2020). Other four sub-branches, including instance re-weighting, classifier adaptation, deep adaptation and adversarial adaptation, are closely connected to the main stream towards universal representation. The milestones and key time nodes of TAL in the past decade have been flagged on each branch.}
\label{fig_framework}
\end{figure*}

\subsection{Open Set Recognition}
Conventional recognition tasks in computer vision where all testing classes are known at training time are generally recognized as \textit{closed-set} recognition. Open set recognition addresses a more realistic vision scenario where unknown classes can be encountered during testing time~\cite{Scheirer2013Open, Scheirer2014Open, Li2005Open, Zhang2016Open}, which shares very similar characteristic with ZSL in tasks. ZSL is different from open set recognition that the former uses the semantic embedding of visual features for recognizing unknown classes, while the latter focus on a one-class classification problem.

More recently, a similar open set framework with transductive ZSL for recognition under domain shift is the \textit{open-set} domain adaptation (OSDA) approach~\cite{Gall2016Open, Busto2018Open,PanY2020CVPR,Kundu2020CVPR}, which were established on the concept of open set recognition. Conventional domain adaptation assumes that the categories in target domain are known and can be seen in the source domain, while open-set domain adaptation addresses the scenarios where the target domain contains the instances of categories that are unseen in the source domain~\cite{Busto2018Open}. The differences between zero-shot learning and open-set domain adaptation lie in that (1) \textit{Conventional} ZSL tends to solve the recognition of instances of unseen categories under the \textit{same} marginal distribution across training and testing data, while open set domain adaptation aims to solve the same problem but under \textit{different} marginal distribution across source and target domains. (2) \textit{Generalized} ZSL~\cite{Rahman2018ZSL, Verma2018GZSL} was proposed for the scenario where the training and test classes are not necessarily disjoint, while open set domain adaptation was proposed for the scenario where there still a few categories of interest are shared across source and target data. The open set domain adaptation shares some similarity with ZSL that some classes are unseen. The difference lies in that the unseen classes in OSDA are directly recognized as \textit{unknown} class.

This paper provides a novel taxonomy of TAL regardless of \textit{closed-set}, \textit{open-set} and \textit{partial-set} split.
In the past 30 years, the overall research context of transfer adaptation learning (TAL), including the origin of transfer concept, establishment of domain adaptation theory, and rapid development of TAL models and algorithms towards universal representation, is illustrated in Fig.~\ref{fig_framework}. The milestones with respect to those representative breakthroughs in theory and models around the taxonomy have been flagged.


\section{Instance Re-weighting Adaptation}

When the training and test data are drawn from different distribution, this is commonly referred to as \textit{sample selection bias} or \textit{covariate shift}~\cite{Huang2007In}. Instance re-weighting aims to infer the resampling weight directly by feature distribution matching across different domains in a non-parametric manner. Generally, given a dataset $(x,y)\sim P_r(x,y)$, a learning model can be obtained by minimizing the following expected risk of the training set,
\begin{equation}
\begin{split}
 &R\lbrack P_r, \theta, l(x, y, \theta)\rbrack=E_{(x,y)\sim P_r(x,y)}\lbrack l(x, y, \theta) \rbrack\\
\end{split}
\end{equation}

But actually, we are more concerned about the expected risk of the testing set, shown as follows
\begin{equation}
\begin{split}
 &R\lbrack P_r', \theta, l(x, y, \theta)\rbrack=E_{(x,y)\sim P_r'(x,y)}\lbrack l(x, y, \theta) \rbrack\\
 &   =E_{(x,y)\sim P_r(x,y)}\lbrack \frac{P_r'(x,y)}{P_r(x,y)}l(x, y, \theta) \rbrack\\
 &   =E_{(x,y)\sim P_r(x,y)}\lbrack \beta(x,y)l(x, y, \theta) \rbrack\\
\end{split}
\label{eq_ratio}
\end{equation}
where $P_r(x,y)$ and $P_r'(x,y)$ represent the probability distribution of training and testing data, respectively. $l(x,y,\theta)$ is the loss function and $\beta(x,y)$ is the ratio between the two probabilities, which is amount to the weighting coefficient. Obviously, when $P_r(x,y)=P_r'(x,y)$, we have $\beta(x,y)=1$.

From Eq.(\ref{eq_ratio}), we know that $P_r(x,y)$ and $P_r'(x,y)$ can be estimated for computing the weight $\beta(x,y)$ by following~\cite{Zadrozny2004SSB} based on the prior knowledge of the class distributions. Although this is intuitive, it requires very good density estimation of $P_r(x,y)$ and $P_r'(x,y)$. Particularly, a serious overweighting of the observations with very large coefficients $\beta(x,y)$ will be resulted from possible small errors or noise in estimating $P_r(x,y)$ and $P_r'(x,y)$.
Therefore, in order to improve the reliability of the weights, $\beta(x,y)$ can be directly estimated by imposing flexible constraints into the learning model without having to estimate the two probability distributions.

Sample re-weighting based domain adaptation methods mainly focuses on the case where the difference between the source domain and the target domain is not too large. The objective is to re-weight the source samples so that the source data distribution can be more close to the target data distribution. Usually, when the distribution difference between the two domains is relatively large, the sample re-weighting methods can be combined with others (e.g. feature adaptation) for auxiliary transfer learning. Instance re-weighting has been studied with different models, which can be divided into three categories based on weighting scheme: (i) \textit{Intuitive Weighting}, (ii) \textit{Kernel Mapping Based Weighting}, and (iii) \textit{Co-training Based Weighting}. This kind of methods put emphasis on the learning or computation of the weights by using different criterions and training protocols. The taxonomy of instance re-weighting based models is summarized in Table~\ref{tab1}.

\ifCLASSOPTIONcaptionsoff
  \newpage
\fi

\begin{table}[htbp]
\centering
\caption{Our Taxonomy of Instance Re-weighting Adaptation Approaches}
\begin{tabular}{cccccc}
 \hline
 \scshape \thead{Re-weighting Adaptation} & \scshape Model Basis & \scshape Reference \\
 \hline
 \textbf{\thead{Intuitive Weighting}} & Adaptive tuning & \thead{\cite{Jiang2007Weig, Wang2017Weig},\\\cite{Dai2007Boosting, S.Chen2017Visual}} \\
 \hline
 \textbf{Kernel Map-Based} &  &  \\
 \hline
 Distribution Matching & KMM\&MMD & \thead{\cite{Huang2007In,Chu2013Selective},\\\cite{Yan2017WMMD}} \\
 \hline
 Sample Selection & \thead{K-Means\\\&$l_{21}$-norm} & \cite{Zhong2009Cross, Long2014Transfer} \\
 \hline
 \textbf{Co-training-Based} & Double classifiers & \cite{Chen2011Co,Chen2018RAAN}\\
 \hline
\end{tabular}
\label{tab1}
\end{table}

\subsection{Intuitive Weighting}

Instance re-weighting based domain adaptation was first proposed for natural language processing (NLP)~\cite{Jiang2007Weig, Wang2017Weig}. In~\cite{Jiang2007Weig}, Jiang and Zhai proposed an intuitive instance weighted domain adaptation framework, which introduced four parameters for characterizing the distribution difference between source and target samples. For example, for each $(x_i^{s},y_i^{s}) \in \mathcal{D}_s$, the labeled source data, the parameter $\alpha_i$ that was used to indicate how likely $\mathcal{P}_{target}(y_i^s\vert x_i^s)$ is close to $\mathcal{P}_{source}(y_i^s\vert x_i^s)$ and the parameter $\beta_i$ that was ideally computed as $\frac{\mathcal{P}_{target}(x_i^s)}{\mathcal{P}_{source}(x_i^s)}$ were introduced. Obviously, large $\alpha_i$ means the high confidence of the labeled source sample $(x_i^{s},y_i^{s})$ contributing positively to the learning effectiveness. Small $\alpha_i$ means the two probabilities are very different, and the instance $(x_i^{s},y_i^{s})$ can be \textit{discarded} in the learning process. Additionally, for each $x_i^{t,u} \in \mathcal{D}_{t,u}$, the unlabeled target data, and for each possible label $y \in \mathcal{Y}$, the hypothesis space, the parameter $\gamma_i(y)$ that indicates how likely a tentative pseudo-label $y$ can be assigned to $x_i^{t,u}$, then the $(x_i^{t,u},y)$ is \textit{included} as a training sample.

Generally, $\alpha_i$ and $\gamma_i$ play an intuitive role in sample selection by removing those misleading source samples and adding those valuable labeled target samples during the transfer learning process. Although the optimal weighting values of these parameters for the target domain are unknown, the intuitions behind the weights can be served as guidelines for researchers designing heuristic parameter tuning scheme~\cite{Jiang2007Weig}. Therefore, adaptive learning of these intuitive weights remains still a challenging issue.

In~\cite{Wang2017Weig}, Wang et al. proposed two instance weighting schemes for neural machine translation (NMT) domain adaptation, i.e., sentence weighting and dynamic domain weighting. Specifically, given the parallel training corpus $\mathcal{D}=\lbrack \mathcal{D}_{in},\mathcal{D}_{o}\rbrack$ consisting of in-domain data and out-of-domain data, the sentence weighted NMT objective function was written as
\begin{equation}
\begin{split}
 &\mathcal{J}_{sw}=\sum_{\langle x_i,y_i\rangle\in \mathcal{D}}\lambda_i\log\mathcal{P}(y_i\vert x_i)\\
\end{split}
\label{sentence_weigh}
\end{equation}
where $\lambda_i$ is the weight to score each $\langle x_i,y_i\rangle$. $\mathcal{P}(\cdot)$ is the conditional probability activated by softmax function. $x$ and $y$ represent the source sentence and target sentence, respectively.
For domain weighting (dw), a weight $\lambda$ was designed for the in-domain data, and the NMT objective function Eq.(\ref{sentence_weigh}) can be transformed as~\cite{Wang2017Weig}
\begin{equation}
\begin{split}
 &\mathcal{J}_{dw}=\lambda\sum_{\langle x,y\rangle\in\mathcal{D}_{in}}\log\mathcal{P}(y\vert x)+\sum_{\langle x',y'\rangle\in\mathcal{D}_{o}}\log\mathcal{P}(y'\vert x')\\
\end{split}
\end{equation}
A dynamic batch weight tuning scheme was proposed by monotonically increasing the ratio of in-domain data in the minibatch, which is supervised by the training cost. Dai et al. proposed a TrAdaBoost~\cite{Dai2007Boosting} transfer learning method, which leveraged Boosting algorithm to automatically tune the weights of the training samples.

In~\cite{S.Chen2017Visual}, Chen et al. proposed a more intuitive weighting based subspace alignment method by re-weighting the source samples for generating source subspace that are close to the target subspace. Let $w=\lbrack w_1,\cdots, w_m\rbrack^T\in \mathcal{R}^m$ denote the weighting vector of the source samples. Obviously, the $w_i$ w.r.t. the source sample $x_i$ increases if its distribution is more close to target data. Therefore, a simple weight assignment strategy was presented for assigning larger weights to the source samples that are closer to target domain~\cite{S.Chen2017Visual}.

After obtaining the weight vector $w$, the weighted source space can be obtained by performing PCA on the following covariance matrix $\mathcal{C}$ of weighted source data,
\begin{equation}
\begin{split}
 &\mathcal{C}=\frac{1}{m}\sum_{i=1}^m(x_i-\mu)^Tw_i(x_i-\mu)\\
\end{split}
\end{equation}
where $\mu$ is the weighted mean vector. Then the eigenvectors $P_S$ can span the source subspace. By performing PCA on the target data, the eigenvectors $P_T$ can span the target subspace. Thereafter, the following unsupervised domain adaptation model, \textit{subspace alignment} (SA)~\cite{Fernando2013SA}, with Frobenius norm minimization, was implemented.
\begin{equation}
\begin{split}
 &\min_M\Arrowvert P_SM-P_T\Arrowvert_F^2\\
\end{split}
\label{SA_model}
\end{equation}
The subspace alignment matrix $M$ can be easily solved with least-square solution.

\subsection{Kernel Mapping Based Weighting}
The intuitive weighting based domain adaptation was implemented in the raw data space. In order to infer the sampling weights by distribution matching across source and target data in feature space in a non-parametric way, kernel mapping based weighting was proposed. Briefly, the distribution difference between source and target data can be better characterised by re-weighting the source samples such that the means of the source and target instances in a reproducing kernel Hilbert space (RKHS) are close~\cite{Huang2007In}. Kernel mapping based weighting consists of two categories of methods: \textit{Distribution Matching}~\cite{Huang2007In, Chu2013Selective, Yan2017WMMD} and \textit{Sample Selection}~\cite{Zhong2009Cross, Long2014Transfer}.

\textit{(1) Distribution Matching}. The intuitive idea of distribution matching is to match the means between the source and target data in a reproducing kernel Hilbert space (RKHS) by resampling the weights of the source data. Two similar distribution matching criterions, i.e., kernel mean matching (KMM)~\cite{Huang2007In} and maximum mean discrepancy (MMD)~\cite{Gretton2006MMD, Gretton2012MMD}, have been used as non-parametric statistic to measure the distribution difference. Specifically, Huang et al.~\cite{Huang2007In} firstly proposed to re-weight the source samples with $\beta$, such that the KMM between the means of target data and the weighted source data is minimized.
\begin{equation}
\begin{split}
 &\min_{\beta}\Arrowvert E_{x'\sim P_r'}\lbrack\Phi(x')\rbrack-E_{x\sim P_r}\lbrack\beta(x)\Phi(x)\rbrack\Arrowvert\\
 &s.t.\quad \beta(x)\geq0, E_{x\sim P_r}\lbrack\beta(x)\rbrack=1\\
\end{split}
\end{equation}
where $\Phi(\cdot)$ is the nonlinear mapping function into RKHS.

Chu et al.~\cite{Chu2013Selective} further proposed a selective transfer machine (STM) by minimizing KMM for distribution matching, and simultaneously minimizing the empirical risk of the classifiers learned on the reweighted training samples.
\begin{equation}
\begin{split}
 &(w,s)= \arg\min_{w,s} R_w(\mathcal{D}^{tr},s)+\lambda \Omega_s(\mathcal{D}^{tr},\mathcal{D}^{te}) \\
\end{split}
\label{kmm_reg}
\end{equation}
where $R_w(\cdot)$ is the empirical risk (loss) on the training set $\mathcal{D}^{tr}$, $\Omega_s$ indicates the distribution mismatch formulated by KMM, $s$ is the weighting vector of the source samples, and $w$ is the classifier parameters. From Eq.(\ref{kmm_reg}), the KMM based distribution mismatch plays an important role in model regularization on the sampling weights.

More recently, Yan et al.~\cite{Yan2017WMMD} proposed a weighted MMD (WMMD) for domain adaptation, which was implemented with convolutional neural network. WMMD overcomes the flaw of conventional MMD that ignores the class weight bias and assumes the same class weights between source and target domain. WMMD is formulated as~\cite{Yan2017WMMD}
\begin{equation}
\begin{split}
 &d_{wmmd}^2= \Arrowvert\frac{1}{\Sigma_{i=1}^M\alpha_{y_i^s}}\sum_{i=1}^M\alpha_{y_i^s}\phi(x_i^s)-\frac{1}{N}\sum_{j=1}^N\phi(x_j^t) \Arrowvert_\mathcal{H}^2\\
\end{split}
\end{equation}
where $\alpha_{y_i^s}$ is the class weight w.r.t. the class $y_i^s$ of the $i^{th}$ source sample and $\phi(\cdot)$ is the nonlinear mapping into RHKS $\mathcal{H}$. $M$ and $N$ denote the number of samples drawn from source and target domain, respectively.

\textit{(2) Sample Selection} is another kind of kernel mapping based re-weighting method. Zhong et al.~\cite{Zhong2009Cross} proposed a cluster based sample selection method KMapWeighted which was established on the assumption that the kernel mapping can make the marginal distribution across domains similar, but the conditional probabilities between two domains after kernel mapping are still different. Therefore, in the RKHS space, they further select those source samples that are more likely similar to target data via a $K$-means based clustering criterion. The data in the same cluster should be with the same labels and then the source samples with similar labels to target data were selected.

Long et al.~\cite{Long2014Transfer} proposed a TJM method for domain adaptation by minimizing the MMD based distribution mismatch between source and target data, in which the transformation matrix $A$ was imposed with structural sparsity (i.e., $l_{2,1}$-norm regularization constraint) for sampling. Then, larger coefficients correspond to the strong correlation between the source samples and the target domain samples. The TJM model is provided as~\cite{Long2014Transfer}
\begin{equation}
\begin{split}
 &\min_{A^TMA=I}tr(A^TMA)+\lambda(\Arrowvert A_s\Arrowvert_{2,1}+\Arrowvert A_t\Arrowvert_F^2)\\
\end{split}
\end{equation}
where the $l_{2,1}$-norm on source transformation $A_s$ means that source outliers can be excluded in transferring to target domain, the target transformation $A_t$ was regularized for smoothness, and $M=KHK^T$ is the deduced matrix from MMD. $H$ is the centering matrix and $K$ is kernel matrix.
\subsection{Co-training Based Weighting}
Co-training~\cite{Blum2005Co} assumes that the dataset is characterized into two different views, in which two classifiers are then separately learned for each view. The inputs with high confidence of one of the two classifiers can be moved to the training set. In weighting based transfer learning, Chen et al. proposed a CODA~\cite{Chen2011Co} method, in which two classifiers with different weight vectors were trained. For better training both classifiers on the training set, the two classifiers were jointly minimized with weighting. In essence, the method of sample re-weighting based on the classifier is similar to the TrAdaBoost~\cite{Dai2007Boosting} and KMapWeighted~\cite{Zhong2009Cross}.

In~\cite{Chen2018RAAN}, Chen et al. proposed a re-weighted adversarial adaptation network (RAAN) for unsupervised domain adaptation. Two classifiers including a multi-class source instance classifier $\mathcal{C}$ and a binary domain classifier $\mathcal{D}$ were designed for adversarial training. The domain classifier $\mathcal{D}$ aims to discriminate whether features are from source or target domain, while the domain feature representation network $\mathcal{T}$ tries to confuse them, which formulates an adversarial training manner. For improving the domain confusion effect, the source feature distribution is re-weighted with $\beta$ during training of the domain classifier $\mathcal{D}$. With the gaming between $\mathcal{T}$ and $\mathcal{D}$ as GAN does~\cite{Goodfellow2014GAN}, the following minimax objective function was used~\cite{Chen2018RAAN},
\begin{equation}
\begin{split}
 &\min_{\mathcal{T}}\max_{\mathcal{D}, \beta}\mathcal{L}_{adv}^{Re}\\
\end{split}
\end{equation}
where the weight $\beta$ is multiplied with $\mathcal{D}$, and both $\beta$ and $\mathcal{D}$ were trained in a cooperative way. The learning of the source classifier $\mathcal{C}$ was easily performed by minimizing the cross-entropy loss.

\subsection{Discussion and Summary}
In this section, we recognize three kinds of instance re-weighting: intuitive, kernel mapping and co-training. The intuitive re-weighting advocates to tune the weights of the source samples, such that the weighted source distribution is closer to target distribution. The kernel mapping based re-weighting is further divided into distribution matching and sample selection. The former aims to learn source sample weights such that the kernel mean discrepancy between target data and the weighted source data is minimized, and the latter advocates sample selection by using K-means clustering (cluster assumption) and $l_{2,1}$-norm based structural sparsity in RKHS space. The co-training mechanism focuses on learning with two classifiers. Additionally, the adversarial training of the weighted domain classifier can facilitate domain confusion.

Although instance re-weighting is the earliest method to address domain mismatch problem, there are still some directions worth studying: 1) essentially, instance weighting can be incorporated into most of learning frameworks; 2) the initialization and estimation of instance weights are important and can be treated as a latent variable obeying some probability distribution.

\section{Feature Adaptation}

Feature adaptation aims to discover the common feature representation of the data drawn from multiple sources by using different techniques including linear and nonlinear ones. In the past decade, feature adaptation induced transfer adaptation learning has been intensively studied, which, in our taxonomy, can be categorized into (i) \textit{Feature Subspace}-Based, (ii) \textit{Feature Transformation}-Based, (iii) \textit{Feature Reconstruction}-Based and (iv) \textit{Feature Coding}-Based. Despite these advances, the technical challenges being faced by researchers lie in the domain subspace alignment, projection learning for distribution matching, generic representation and shared domain dictionary coding. The taxonomy of feature adaptation approaches is summarized in Table~\ref{tab2}.

\begin{table}[htbp]
\centering
\caption{Our Taxonomy of Feature Adaptation Approaches}
\begin{tabular}{cccccc}
 \hline
 \scshape Feature Adaptation & \scshape Model Basis & \scshape Reference \\
 \hline
 \textbf{Feature Subspace} &  &  \\
 \hline
 Geodesic path & Grassman manifold & \cite{Gopalan2011SGF,GongShiShaEtAl2012}\\
 \hline
 Alignment & Subspace learning & \thead{\cite{Fernando2013SA,Sun2015Subspace},\\\cite{Liu2018GTH,FuJR2020TNNLS}}\\
 \hline
 \textbf{Feature Transformation} &  &  \\
 \hline
 Projection & \thead{MMD\&HSIC\&\\Bregman divergence} & \thead{\cite{PanTsangKwokEtAl2011,LongWangDingEtAl2014},\\\cite{Xiao2015Kernel,SiTaoGeng2010}} \\
 \hline
 Metric & \thead{First/second-order \\statistic} & \thead{\cite{Z.Ding2017Robust, H.Wang2014Cross},\\\cite{Sun2016DeepCoral, Herath2017ILS}} \\
 \hline
 Augmentation & \thead{Zero-padding\&\\Generative} & \thead{\cite{Daume2010Frust,Li2014Augment},\\\cite{Volpi2018Augment,Zhang2017MCTL}}\\
 \hline
 \textbf{Feature Reconstruction} &  & \\
 \hline
 Low-rank models& \thead{Low-rank \\representation (LRR)} & \thead{\cite{JhuoLiuLeeEtAl2012,ShaoKitFu2014},\\\cite{Zhang2017MCTL,Fu2018GSL}}\\
 \hline
 Sparse models& \thead{Sparse subspace \\clustering (SSC)} & \thead{\cite{Zhang2016LSDT,XuFangWuEtAl2016},\\\cite{Zhang2017DKTL,Wang2017CRTL}}\\
 \hline
 \textbf{Feature Coding} & & \\
 \hline
 Domain-shared dictionary & Dictionary learning &\thead{\cite{Shekhar2013SDDL,Xu2015DADL},\\\cite{Shekhar2015CPAD, Qiu2015CDDA}}\\
 \hline
 Domain-specific dictionary & Dictionary learning &\thead{\cite{Qiu2012DADL, Ni2013SIDL},\\\cite{Zhu2014CDDL, Lu2015SIDL}}\\
 \hline
\end{tabular}
\label{tab2}
\end{table}

\subsection{Feature Subspace-Based}
Learning subspace generally resorts to unsupervised domain adaptation. Three representative models are referred to as sampling geodesic flow (SGF)~\cite{Gopalan2011SGF}, geodesic flow kernel (GFK)~\cite{GongShiShaEtAl2012} and subspace alignment (SA)~\cite{Fernando2013SA}. There exists a common property of the three methods, i.e. the data is assumed to be represented by a low-dimensional linear subspace. That is, a low-dimensional Grassmann manifold is embedded in the high-dimensional data. Generally, principal component analysis (PCA) was used to construct the Grassmann manifold, where the source and target domains become two points and a geodesic flow or path was formulated.
SGF proposed by Gopalan, et al.~\cite{Gopalan2011SGF} is an unsupervised low-dimensional subspace transfer method, which samples a group of subspaces along the geodesic path between source and target data, and aims to find an intermediate representation with closer domain distance.

Similar to but different from SGF, Gong et al. proposed a GFK~\cite{GongShiShaEtAl2012}, in which the geodesic flow kernel was used to model the domain shift by integrating an infinite number of subspaces. GFK explores an intrinsic low-dimensional spatial structure that associates two domains and the main idea behind is to find a geodesic line from $\phi(0)$ to $\phi(1)$, such that the raw feature can be transformed into a space of infinite dimension from $\phi(0)$ to $\phi(1)$ where distribution difference is easy to be reduced. In particular, the infinite dimensional features in the manifold space can be represented as $z = {\phi(t)}^Tx$. The inner product of the transformed features $z_i$ and $z_j$ defines a positive semi-definite geodesic flow kernel as follows:
\begin{equation}
\begin{split}
 & \biggl< z_i^\infty ,z_i^\infty \biggr> = \int_0^1 {({\phi(t)}^T x_i)}^T ({\phi(t)}^T x_i) dt = x_i^TGx_j\\
\end{split}
\end{equation}
where $G$ is a positive semi-definite mapping matrix. With $z=\sqrt G$x, features in the original space can be transformed into the Grassmann manifold space.

For aligning the source subspace to the target subspace, in SA~\cite{Fernando2013SA}, Fernando, et al. proposed to move closer the two subspaces with respect to the points in Grassmann manifold by directly designing an alignment matrix $M$, which well bridges the source and target subspaces. The model of SA is described in Eq.(\ref{SA_model}). As presented in SA, the subspaces of source and target data were spanned by the eigenvectors induced with a PCA. Further, Sun and Saenko proposed a subspace distribution alignment (SDA)~\cite{Sun2015Subspace} by simultaneously aligning the distributions as well as the subspace bases, which overcomes the flaw of SA that does not take into account the distribution difference.

More intuitively, Liu and Zhang proposed a guided transfer hashing (GTH)~\cite{Liu2018GTH} framework, which introduced a more generic method for moving the source subspace $W_s$ closer to target subspace $W_t$,
\begin{equation}
\begin{split}
 & \min_{W_s, W_t}\frac{1}{2}\Arrowvert M^{\frac{1}{2}}\odot(W_t-W_s)\Arrowvert^2\\
\end{split}
\label{sub_close}
\end{equation}
where $M$ is a weighting matrix on the difference between source and target subspaces. Through this way, the two subspaces can be solved alternatively and progressively, which is therefore recognized as a guided transfer mechanism. Further, a guide subspace learning (GSL) was proposed for unsupervised domain adaptation~\cite{FuJR2020TNNLS}.

\subsection{Feature Transformation-Based}
This kind of models aim to learn a transformation or projection of the data with some distribution matching metrics between source and target domains~\cite{Saenko2010symm, Kulis2015Art,Courty2017Transport}. Then, the transformed or projected feature distribution difference across two domains can be removed or relieved. Feature transformation based domain adaptation has been a main-stream in visual transfer learning community in last years, which can be further divided into \textit{Projection}, \textit{Metric}, and \textit{Augmentation} according to the model formulation.

\textit{(1) Projection-Based} domain adaptation aims to solve a projection matrix in source and target domain for reducing the marginal distribution difference and conditional distribution difference between domains, by introducing \textit{Kernel Matching Criterion}~\cite{SiTaoGeng2010,PanTsangKwokEtAl2011, Baktashmotlagh_2013_ICCV, LongWangDingEtAl2014, Xiao2015Kernel, Long2015DITKL, Zhang2017DRCA, Ghifary2017SCA} and \textit{Discriminative Criterion}~\cite{Zhang2017Joint, Zhang2017CDSL, Lu2018An,Shuang.Li2018Domain}. The kernel matching criterion generally adopts the maximum mean discrepancy (MMD) statistic, which characterizes the \textit{marginal} distribution difference and \textit{conditional} distribution difference between source and target data. In unsupervised domain adaptation setting, the labels of target domain samples are generally unavailable, therefore the pseudo-labels of target samples should be iteratively predicted for quantifying the conditional MMD between domains~\cite{Zhang2017Joint, Li2018CDELM}. The discriminative criterion focus on within-class compactness and between-class separability of the projection. Mathematically, the formulation of empirical nonparametric MMD in universal RKHS is written as
\begin{equation}
\begin{split}
 &d_{\mathcal{H}}^2(\mathcal{D}_s,\mathcal{D}_t)= \Arrowvert\frac{1}{M}\sum_{i=1}^M\phi(x_i^s)-\frac{1}{N}\sum_{j=1}^N\phi(x_j^t) \Arrowvert_\mathcal{H}^2\\
\end{split}
\label{MMD_fun}
\end{equation}

Specifically, with MMD based kernel matching criterion, Pan and Yang firstly proposed a transfer component analysis (TCA)~\cite{PanTsangKwokEtAl2011} by introducing the marginal MMD with projection as the loss function. The joint distribution adaptation (JDA) proposed by Long et al.~\cite{LongWangDingEtAl2014} further introduced the conditional MMD on the basis of TCA, such that the cross-domain distribution alignment becomes more discriminative. The general model can be written as
\begin{equation}
\begin{split}
 & \min_{W}d_{m}^2(X_S,X_T,W)+\lambda d_{c}^2(X_S,X_T,Y_S,Y_T',W)\\
\end{split}
\end{equation}
where $W$ denotes the projection matrix, $Y_T'$ denotes the predicted pseudo-label of target data, $d_m^2$ and $d_c^2$ represent the marginal and conditional distribution discrepancy, respectively.
For improving the discrimination of the projection matrix, such that the within-class compactness and between-class separability in each domain can be better characterized, the model with joint discriminative subspace learning and MMD minimization was proposed, for example, JGSA~\cite{Zhang2017Joint} and CDSL~\cite{Zhang2017CDSL}, and generally written as
\begin{equation}
\begin{split}
 & \min_{W}F(W, X_S, X_T, Y_S, Y_T')+\lambda d_{\{m,c\}}^2(X_S,X_T,W)\\
\end{split}
\label{SubFun}
\end{equation}
where $F(\cdot)$ is a scalable subspace learning function of the projection $W$, for example, linear discriminative analysis (LDA), local preservation projection (LPP), marginal fisher analysis (MFA), principal component analysis (PCA), etc.
In addition to the MMD based criterion in projection based transfer model, Bregman divergence based~\cite{SiTaoGeng2010}, Hilbert-Schmidt independence criterion (HSIC) based~\cite{Gretton2005HSIC, Xiao2015Kernel, Yan2017HSIC, Wang2017CRTL}, and manifold criterion based~\cite{Zhang2017MCTL}.

In~\cite{SiTaoGeng2010}, Si et al. proposed a transfer subspace learning (TSL) by introducing a Bregman divergence-based discrepancy as regularization instead of MMD, which is written as
\begin{equation}
\begin{split}
 & W=arg \min_{W} F(W) +\lambda D_W(P_L||P_U)\\
\end{split}
\label{SubFun1}
\end{equation}
where $F(W)$ is similar to Eq.(\ref{SubFun}) and $D_W(P_L||P_U)$ is the Bregman divergence-based regularization that measures the distance between the probability distribution of training samples $P_L$ and that of the testing samples $P_U$ in the projected subspace $W$.

The HSIC proposed by Gretton et al.~\cite{Gretton2005HSIC}, the same author as that of MMD, was used to measure the dependency between two sets $\mathcal{X}$ and $\mathcal{Y}$. Let $k_x$ and $k_y$ denote the kernel function w.r.t. the RKHS $\mathcal{F}$ and $\mathcal{G}$. The HSIC is mathematically written as~\cite{Gretton2005HSIC}
\begin{equation}
\begin{split}
 & HSIC(\mathcal{X},\mathcal{Y},\mathcal{F},\mathcal{G})\\
 &=\Arrowvert C_{\mathcal{XY}}\Arrowvert_{H-S}^2
  =(N-1)^{-2}Tr(\mathcal{K_X}H\mathcal{K_Y}H)\\
 & s.t.\quad H=I-N^{-1}1_{N\times1}1_{N\times1}^T\\
\end{split}
\end{equation}
where $N$ is the size of the set $\mathcal{X}$ and $\mathcal{Y}$ and $\Arrowvert C_{\mathcal{XY}}\Arrowvert_{H-S}^2$ is the Hilbert-Schmidt norm of the cross-covariance operator. $\mathcal{K_X}$ and $\mathcal{K_Y}$ denote the two kernel Gram matrix, and $H$ is the centering matrix. HSIC will be zero if and only if $\mathcal{X}$ and $\mathcal{Y}$ are independent. In~\cite{Wang2017CRTL}, Wang et al. proposed to use the projected HSIC as regularization, which is written as
\begin{equation}
\begin{split}
 & \min_{W} F(W) -\lambda HSIC(W,\mathcal{X},\mathcal{Y},\mathcal{F},\mathcal{G})\\
\end{split}
\label{SubFun2}
\end{equation}
where $\mathcal{Y}$ denotes the label set of source and target data. Obviously, the model constrains $W$ to reduce the independency between feature set $\mathcal{X}$ and label set $\mathcal{Y}$, such that the classification performance can be improved. In model formulation, the general way is to set a common projection for both domains. Another way is to learn two projections $W_S$ and $W_T$, one for each domain, such that domain specific projection can be solved~\cite{Zhang2017Joint, Herath2017ILS, Liu2018GTH,Fu2018GSL}. For moving the two projections of both domains closer, the Frobenius norm of their difference like Eq.(\ref{sub_close}) can be used.

\textit{(2) Metric-Based} aims to learn a good distance metric from labeled source data which can be easily adapted to a related but different target domain~\cite{Zhang2010TML}. Metric transfer has a close link to projection based examples, if the metric $M$ is a semi-definite matrix and can be decomposed into $M=WW^T$~\cite{Z.Ding2017Robust}. The metric-based transfer can be divided into \textit{First-order statistic}~\cite{Z.Ding2017Robust, H.Wang2014Cross, Kulis2015Art, Junlin2016Deep, Geng2011DAML, Xu2017MTL,LiS2020TNNLS} and \textit{Second-order statistic}~\cite{Sun2016DeepCoral, Herath2017ILS, SunFengSaenko2017,Zhen2018Aligning,Li2018DACM} based distance metric, such as Euclidean, Mahalanobis distance, moment distance, etc.

\textit{The First-order} metric transfer generally learns a metric under which the distance between source and target feature is minimized, and it can be written as
\begin{equation}
\begin{split}
 & \min_{M} d(M,\phi(X_S),\phi(X_T)) + \lambda\Re(M)\\
\end{split}
\end{equation}
where $\phi(\cdot)$ is the feature representation or mapping function, and it can be linear mapping~\cite{Kulis2015Art}, kernel mapping~\cite{H.Wang2014Cross,Geng2011DAML}, auto-encoder~\cite{Z.Ding2017Robust} or neural network~\cite{Junlin2016Deep}.

For example, the robust transfer metric learning (RTML) proposed by Ding et al.~\cite{Z.Ding2017Robust} adopted an auto-encoder based feature representation for metric learning, such that the Mahalanobis distance between source and target domain is minimized. The objective function of RTML is as follows:
\begin{equation}
\begin{split}
 & \min_{M\in S_+^d} \sum_{i=0}^c tr(\phi_iM)+ \alpha {\begin{Vmatrix} \overline X-M \widetilde{X} \end{Vmatrix}}_F^2+\lambda rank(M)\\
\end{split}
\end{equation}
where $M$ is positive semi-definite matrix, $\overline X$ is the repeated version of $X$, $\widetilde{X}$ is the randomly corrupted version of $\overline X$. The first item is Mahalanobis distance induced domain discrepancy under metric $M$, the second item is auto-encoder for feature learning, and the third term is the low-rank constraint for characterizing the internal correlation between domains.

\textit{The Second-order} metric transfer generally learns a metric under which the distance between the covariances of source and target domain instead of the means is minimized~\cite{Zhen2018Aligning, Herath2017ILS, SunFengSaenko2017, Sun2016DeepCoral}. For example, Sun et al.~\cite{SunFengSaenko2017, Sun2016DeepCoral} proposed a simple but efficient correlation alignment (CORAL) by aligning the second-order statistic (i.e. the covariance) between source and target distributions instead of the first-order metric. By introducing a metric matrix $A$, the difference between source covariance $\Sigma_S$ and target covariance $\Sigma_T$ in CORAL can be minimized by solving
\begin{equation}
\begin{split}
 & \min_A \Arrowvert A^T\Sigma_SA-\Sigma_T\Arrowvert_F^2\\
\end{split}
\label{CovFun}
\end{equation}
The Eq.(\ref{CovFun}) is amount to matching the two centered Gaussian distribution, which is the basic assumption for such second-order statistic based transfer.

\textit{(3) Augmentation-Based} domain adaptation often assume that the feature representation is grouped with three types: common representation, source-specific representation and target-specific representation. In general case, the source domain should be characterized as the composition of common component and source-specific component, and similarly, the target domain should be characterized as the composition of common component and target-specific component. Feature augmentation based DA can be divided into the generic \textit{Zero Padding}~\cite{Daume2009Frust,Daume2010Frust,Daume2010CoDA,Li2014Augment,Chen2018ReID} and the latest \textit{Generative}~\cite{Volpi2018Augment,Zhang2017MCTL} types.

\textit{Zero Padding} was firstly proposed by Daume III~\cite{Daume2009Frust}, which presented an EasyAdapt (EA) model. Assume the raw input data space to be $\mathcal{X}\in\Re^F$, then the augmented feature spaces should be $\mathcal{Y}\in\Re^{3F}$. By defining the mapping functions of source and target domain from $\mathcal{X}$ to $\mathcal{Y}$ as $\Phi_s(\cdot)$ and $\Phi_t(\cdot)$, respectively. Then, there is
\begin{equation}
\begin{split}
&\Phi_s(x)=[x,x,0], \Phi_t(x)=[x,0,x]\\
\end{split}
\label{aug_vector}
\end{equation}
where $0\in\Re^F$ is a zero vector. The first, second and third bits of the augmented feature $\Phi(x)$ in Eq.(\ref{aug_vector}) represent the common, source-specific and target-specific feature component, respectively. However, in heterogeneous domain adaptation that addressing different feature dimensions between source and target domain~\cite{Shi2010Hetero, Zhou2014HHTL,Yan2018Hetero}, for example, cross-modal learning (e.g., images vs. text), Li et al.~\cite{Li2014Augment} argued that such simple zero-padding for dimensionality consistence between domains is not meaningful. The reason is that there would be no correspondences between the heterogeneous features. Therefore, Li et al.~\cite{Li2014Augment} proposed a heterogeneous feature augmentation (HFA) model, which incorporates the projected features together with the raw features for feature augmentation by introducing two projection matrices $P\in\Re^{d_c\times d_s}$ and $Q\in\Re^{d_c\times d_t}$. The augmented feature for source and target domain can be written as
\begin{equation}
\begin{split}
&\Phi_s(x_s)=[Px_s,x_s,0_{d_s}], \Phi_t(x_t)=[Qx_t,0_{d_t},x_t]\\
\end{split}
\end{equation}
where $d_s$ and $d_t$ represent the dimensionality of source and target data, respectively.
For incorporating the unlabeled target data, Daume III further proposed an EA++ model with zero padding based feature augmentation for semi-supervised domain adaptation~\cite{Daume2010Frust,Daume2010CoDA}. Chen et al.~\cite{Chen2018ReID} proposed to use zero padding based camera correlation aware feature augmentation (CRAFT) for cross-view person re-identification.

\textit{Generative} methods used for feature augmentation mainly focus on plausible data generation towards enhancing the robustness of domain transfer. In~\cite{Volpi2018Augment}, Volpi et al. proposed an adversarial feature augmentation by introducing two generative adversarial nets (GANs). The first GAN was used to train the generator $S$ for synthesizing implausible source images (data augmentation) by inputting noise and conditional labels. The second GAN was used to train the shared feature encoder $E$ (feature augmentation) for both domains, by adversarial learning with the synthesized source images via $S$. Finally, the encoder $E$ was used as the domain adapted feature extractor shared by both domains. In~\cite{Zhang2017MCTL}, Zhang et al. proposed a manifold criterion guided intermediate domain generation for feature augmentation, which improved the transfer performance by generating high-quality intermediate features.
\subsection{Feature Reconstruction-Based}
Feature reconstruction between source and target data using a representational matrix for domain transfer has been studied for several years. By linear sample reconstruction in an intermediate representation with low-rankness and sparsity, it can well characterize the intrinsic relatedness and correspondences between source and target domain, while excluding noises and outliers during domain adaptation. To this end, feature reconstruction based domain transfer can be generally divided into three types: \textit{Low-rank Reconstruction}~\cite{JhuoLiuLeeEtAl2012,ShaoKitFu2014,Zhang2017MCTL,Fu2018GSL}, \textit{Sparse Reconstruction}~\cite{Zhang2016LSDT,Zhang2017DKTL,XuFangWuEtAl2016,Wang2017CRTL}, and \textit{Graph Matching}~\cite{Das2018S2S,Das2018hypergraph,Das2018graphmatch}. For the first one, for characterizing the domain differences and uncovering the domain noises, the reconstruction matrix was imposed with low-rank constraint, such that the relatedness between domains can be discovered. For the second one, sparsity or structural sparsity was generally used for transferrable sample selection. For the last one, the graphs of each domain are matched by a correspondence matrix. Methodologically, reconstruction based domain transfer is closely related to low-rank representation (LRR)~\cite{Liu2010Robust, Liu2012LRRPAMI}, matrix recovery~\cite{Lin2013The, WrightGaneshRaoEtAl2009} and sparse subspace clustering (SSC)~\cite{Elhamifar2009SSC, Elhamifar2013SSC,Soltanolkotabi2014SSC}.

\textit{(1) Low-rank Reconstruction} based domain adaptation was firstly proposed by Jhuo et al.~\cite{JhuoLiuLeeEtAl2012}, in which the $W$ transformed source feature was reconstructed by the target domain with low-rank constraint on the reconstruction matrix and $l_{2,1}$-norm constraint on the error.
\begin{equation}
\begin{split}
&\min_{W,Z,E}rank(Z)+\alpha\Arrowvert E\Arrowvert_{2,1}\\
&s.t.\quad WX_S=X_TZ+E, WW^T=I\\
\end{split}
\label{Fun_RDALR}
\end{equation}
However, seeking for an alignment between $WX_S$ and $X_T$ may not transfer knowledge directly, due to the out of domain problem of $W$ for unilateral projection.

On the basis of~\cite{JhuoLiuLeeEtAl2012}, Shao et al.~\cite{ShaoKitFu2014} proposed a latent subspace transfer learning (LTSL), which tends to reconstruct the target data by using the source data as basis in a projected latent subspace.
\begin{equation}
\begin{split}
&\min_{W,Z,E}F(W,X_S)+\lambda_1rank(Z)+\alpha\Arrowvert E\Arrowvert_{2,1}\\
&s.t.\quad W^TX_S=W^TX_TZ+E\\
\end{split}
\label{Fun_LTSL}
\end{equation}
where $F(\cdot)$ is a subspace learning function, similar to Eq.(\ref{SubFun}), Eq.(\ref{SubFun1}) and Eq.(\ref{SubFun2}).
By comparing Eq.(\ref{Fun_RDALR}) to Eq.(\ref{Fun_LTSL}), the major difference lies in the latent space learning of $W$ for both domains in LTSL. Both methods, established on LRR, advocated low-rank reconstruction between domains for transfer learning. As demonstrated in~\cite{Liu2012LRRPAMI}, trivial solution may be easily encountered when handling disjoint subspaces and insufficient data using LRR and a strong independent subspace assumption is necessary.

\textit{(2) Sparse Reconstruction} based domain transfer was established on the SSC, which, different from LRR, is well supported by theoretical analysis and experiments when handling the data near the intersections of subspaces~\cite{Elhamifar2013SSC}. Therefore, in~\cite{Zhang2016LSDT}, Zhang et al. proposed a latent sparse domain transfer (LSDT) model, which jointly learn the sparse coding $Z$ between domains and the latent subspace $W$.
\begin{equation}
\begin{split}
 & \min_{Z,W} {\begin{Vmatrix} Z \end{Vmatrix}}_1 + \lambda_1 {\begin{Vmatrix} WX_T-WXZ \end{Vmatrix}}_F^2\\
 & \quad\qquad\qquad+\lambda_2 {\begin{Vmatrix} X-W^TWX \end{Vmatrix}}_F^2 \\
 & s.t.\quad WW^T = I, \quad {\textbf{1}}_{N_S+N_T}^TZ = {\textbf{1}}_{N_T}^T, \quad Z_{N_S+i,i} = 0,\\
 & \qquad \forall i = 1,...,N_T
\end{split}
\end{equation}
where $X$ is the feature set grouped by $X_S$ and $X_T$.

With the sparsity constraint on $Z$, the most transferrable samples can be selected during domain adaptation, which is more robust to noise or outliers drawn from source domain. The model has also been kernerlized by defining the projection $W$ as the linear representation of $X$. The reconstruction is then implemented in a high-dimensional reproducing kernel Hilbert space (RKHS), based on the Representor theorem. In~\cite{Zhang2017DKTL}, Zhang et al. proposed a $l_{2,1}$-norm constraint based reconstruction transfer model with discriminative subspace learning and the domain-class consistency was guaranteed. The joint constraint with low-rankness and sparsity for the reconstruction matrix was proposed in~\cite{XuFangWuEtAl2016}, such that the global and local structures of data can be preserved.

\textit{(3) Graph Matching} based domain adaptation focused on the graph constructions of both domains with adjacency matrices and solved the correspondence matrix $C$ between graphs for inherent domain similarity matching. In~\cite{Das2018S2S}, the first-order (data matrices) and second-order (adjacency matrices) based matching were formulated.
\begin{equation}
\begin{split}
 & \min_{C} \Arrowvert CX_T-X_S\Arrowvert_F^2 + \lambda\Arrowvert CD_T-D_SC\Arrowvert_F^2+\gamma\mathcal{R}(C)\\
\end{split}
\end{equation}
where $D_S$ and $D_T$ represent the adjacency matrices of graphs for source and target domain, respectively. $\mathcal{R}(C)$ denotes the class regularization.
\subsection{Feature Coding-Based}
In feature reconstruction based transfer models, the focus is the learning of reconstruction coefficients across domains, on the basis of the raw feature of source or target data. Different from that, feature coding based transfer learning put emphasis on seeking a group of basis (i.e., dictionary) and representation coefficients in each domain, which was generally called domain adaptive dictionary learning. The typical dictionary learning approach aims to minimize the representation error of the given data set under a sparsity constraint~\cite{Aharon2006KSDV,Mairal2012DL, Jiang2013KSDV}. The cross-domain dictionary learning aims to learn domain adaptive dictionaries without requiring any explicit correspondences between domains, which was generally divided into two types of learning, \textit{domain-shared dictionary}-based~\cite{Shekhar2013SDDL,Xu2015DADL, Shekhar2015CPAD, Qiu2015CDDA} and \textit{domain-specific dictionary}-based~\cite{Qiu2012DADL, Ni2013SIDL, Zhu2014CDDL, Lu2015SIDL, Li2015CPDL}. Obviously, the former resorts to learning one common dictionary for both domains, while the latter contributes to obtain two or more dictionaries for each domain.

\textit{(1) Domain-shared dictionary} aims at representing the source and target domain using a common dictionary. In~\cite{Shekhar2013SDDL,Shekhar2015CPAD}, Shekhar et al. proposed to separately represent the source and target data in a latent subspace with a shared dictionary $D$, which can be written as
\begin{equation}
\begin{split}
 & \min_{D,P,\alpha}\sum_{k\in\{s,t\}}\Arrowvert P_{(k)}X_{(k)}-D\alpha_{(k)}\Arrowvert_F^2+\Re(D,P,\alpha)\\
\end{split}
\end{equation}
where $P$ denotes the latent subspace projection, $\alpha$ denotes the representational coefficients for source data $X_s$ and target data $X_t$ using a shared dictionary $D$, and $\Re(\cdot)$ denotes the regularizer. The shared dictionary $D$ is demonstrated to incorporate the common information from both domains.

\textit{(2) Domain-specific dictionary} tends to learn multiple dictionaries, one for each domain, to represent the data in each domain based on domain specific or common representation coefficients~\cite{Zhu2014CDDL, Li2015CPDL}. The general model can be written as
\begin{equation}
\begin{split}
 & \min_{D,P,\alpha}\sum_{k\in\{s,t\}}\Arrowvert X_{(k)}-D_{(k)}\alpha_{(k)}\Arrowvert_F^2+\Omega(\alpha_s,\alpha_t)\\
\end{split}
\label{dcd_fun}
\end{equation}
where $\Omega(\cdot)$ denotes the difference between representation coefficients of source and target. If $\alpha_s=\alpha_t=\alpha$, then $\Omega(\alpha_s,\alpha_t)=0$ and the model in Eq.(\ref{dcd_fun}) is degenerated as the common representation coefficients based domain adaptive dictionary learning~\cite{Qiu2012DADL}.

In~\cite{Ni2013SIDL, Xu2015DADL,Lu2015SIDL}, a set of intermediate domains that bridge the gap between source and target domains were incorporated as multiple dictionaries $\{D_k\}_{k=1}^{K-1}$, which can progressively capture the intrinsic domain shift between source domain dictionary $D_0$ and target domain dictionary $D_K$. The difference $\bigtriangleup D_k$ between the atoms of adjacent two sub-dictionaries can well characterize the incremental transition and shift between two domains. Actually, this kind of models can be linked with SGF~\cite{Gopalan2011SGF} and GFK~\cite{GongShiShaEtAl2012} by sampling finite or infinite number of intermediate subspaces on the Grassmann manifold for better capturing the intrinsic domain shift.

\subsection{Discussion and Summary}
In this section, feature adaptation methods are presented, including subspace, transformation, reconstruction and coding based types. Feature subspace focuses on the subspace alignment between domains in Grassmann manifold. Feature transformation is further categorized into three subclasses: projection learning with MMD criterion, metric learning with first-order or second-order statistics and augmentation with zero-padding. Feature reconstruction aims to explicitly bridge the source and target data in a latent subspace by low-rank or sparse reconstruction. Besides the data-level matching, higher-order graph matching between the adjacency matrices of graphs for domains is also introduced. Finally, the feature coding focus on domain data representation by learning domain adaptive dictionaries without explicit correspondences between domains.

Feature adaptation has been intensively studied by addressing negative transfer and under-adaptation problems from different perspectives. Two future directions in feature level are specified: 1) more reliable probability distribution similarity metric is needed, except the Gaussian kernel induced MMD; 2) for learning domain-invariant representation, model ensemble of linear and nonlinear ones is desired.

\section{Classifier Adaptation}
In cross-domain visual categorization, classifier adaptation based TAL aims to learn a generic classifier by leveraging labeled samples drawn from source domain and few labeled samples from target domain~\cite{YangYanHauptmann2007,Duan2012Exploiting,Zhang2014DAELM,Royer2015CAPT}. Typical cross-domain classifier adaptation can be divided into (i) \textit{Kernel Classifier}-Based~\cite{YangYanHauptmann2007, Duan2009DTSVM, Duan2010AMKL, Lixin2012Visual,Duan2012Exploiting,Duan2012Domain,Li2017Examplar}, (ii) \textit{Manifold Regularizer}-Based~\cite{M.Longl2014ARTL,LeiZhang2016,Wang2018Visual,Cao2018SDMM,Yao2015SDASL,Zhang2020eccv,LuoYW2020TPAMI} and (iii) \textit{Bayesian Classifier}-Based~\cite{Gonen2014Bayesian,Gonen2014Bayesian1,Liu2015Bayesian, Gholami2017Bayesian, Karbalayghareh2018Bayesian,Perrone2018Bayesian}. The taxonomy of classifier adaptation approaches is summarized in Table~\ref{tab3}.

\begin{table}[htbp]
\centering
\caption{Our Taxonomy of Classifier Adaptation Approaches}
\begin{tabular}{cccccc}
 \hline
 \scshape \thead{Classifier Adaptation} & \scshape Model Basis & \scshape Reference \\
 \hline
 \textbf{\thead{Kernel Classifier}} & SVM\&MKL & \thead{\cite{YangYanHauptmann2007, Duan2010AMKL,Lixin2012Visual},\\\cite{Duan2012Exploiting,Duan2012Domain,Li2017Examplar}} \\
 \hline
 \textbf{Manifold Regularizer} & \thead{Label Propagation\\\&MMD} & \thead{\cite{M.Longl2014ARTL,LeiZhang2016,Wang2018Visual},\\\cite{Cao2018SDMM,Yao2015SDASL,LuoYW2020TPAMI}} \\
 \hline
 \textbf{Bayesian Classifier} & \thead{Probabilistic\\graph models} & \thead{\cite{Gonen2014Bayesian,Gonen2014Bayesian1,Liu2015Bayesian},\\\cite{Gholami2017Bayesian, Karbalayghareh2018Bayesian,Perrone2018Bayesian}}\\
 \hline
\end{tabular}
\label{tab3}
\end{table}

\subsection{Kernel Classifier-Based}
Yang et al.~\cite{YangYanHauptmann2007} firstly proposed an adaptive support vector machine (ASVM) in 2007 for target classifier training, which assumed that there exists a bias $\Delta f(x)$ between source classifier $f^a(x)$ and target classifier $f(x)$. This means that the bias can be added to the source classifier to generate a new decision function, that is adapted to classifying the target data. There is,
\begin{equation}
\begin{split}
 & f(x) = f^a(x) + \Delta f(x) = f^a(x)+w^T \phi (x)\\
\end{split}
\end{equation}
where $w$ is the parameter of the bias function $\Delta f(x)$, which was solved by standard SVM,
\begin{equation}
\begin{split}
 & \min_w\frac{1}{2}\Arrowvert w\Arrowvert^2+C\sum_{i=1}^N\varepsilon_i\\
 &s.t.\quad y_if^a(x_i)+y_iw^T\phi(x_i)\geq1-\varepsilon_i,\quad \varepsilon_i>0\\
\end{split}
\label{asvm_fun}
\end{equation}
In Eq.(\ref{asvm_fun}), $f^a(\cdot)$ was known and trained on labeled source data, $(x_i,y_i)$ are drawn from few labeled target data, and $w$ is the parameter of $\Delta f(\cdot)$ rather than $f(\cdot)$.

More recently, on the basis of ASVM, Duan et al. proposed a series of multiple kernel learning (MKL) based domain transfer classifiers~\cite{Duan2009DTSVM, Duan2010AMKL, Lixin2012Visual, Duan2012Domain}, including AMKL, DTSVM, and DTMKL, in which the kernel function was assumed to be a linear combination of multiple predefined base kernel functions by following the MKL methodology~\cite{Sonnenburg2006MKL,Xu2008MTA}. Additionally, for reducing the domain distribution mismatch, MMD based kernel matching metric $d_{\Bbbk{}}^2(\cdot)$ was jointly minimized with the structural risk based classifiers. The general model of MKL based classifier adaptation can be written as
\begin{equation}
\begin{split}
 & \min_{\Bbbk{},f}R(\Bbbk{},f,X_S,X_T)+\lambda\Omega(d_{\Bbbk{}}^2(X_S,X_T))\\
\end{split}
\label{mkl_fun}
\end{equation}
where $R(\cdot)$ denotes the structural risk on labeled training samples, $f$ is the decision function, $\Omega(\cdot)$ is the monotonic increasing function, $\Bbbk{}=\sum_{m=1}^Md_mk_m$ is a linear combination of a set of base kernels $k_ms$ with $\sum_{m=1}^Md_m=1$ and $d_m\geqslant0$. The structural risk $R(\cdot)$ was generally formulated based on the hinge loss, i.e., $l_\mathcal{\hbar}(t)=\max(0, 1-t)$, as that in SVM. Duan et al.~\cite{Duan2012DDAR} also proposed a domain adaptation machine (DAM), which incorporated SVM hinge loss based structural risk with multiple domain regularizers for target classifier learning. Regularized least-square loss based classifier adaptation can be referred to as~\cite{M.Longl2014ARTL, LeiZhang2016}.

\subsection{Manifold Regularizer-Based}
The manifold assumption in semi-supervised learning means that the the similar samples with small distance in feature space more likely belongs to the same class. By constructing the affinity graph based manifold regularizer, under which, the classifier trained on source data can be more easily adapted to target data through label propagation. Long et al.~\cite{M.Longl2014ARTL} and Cao et al.~\cite{Cao2018SDMM} proposed ARTL and DMM which advocated manifold regularization based structural risk and between-domain MMD minimization for classifier training, structural preservation and domain alignment. In~\cite{Yao2015SDASL}, Yao et al. proposed to simultaneously minimize the classification error, preserve the geometric structure of data and restrict similarity characterized on unlabeled target data. Zhang and Zhang~\cite{LeiZhang2016} proposed a manifold regularization based least-square classifier EDA on both domains with label pre-computation and refining for domain adaptation. More recently, Wang et al.~\cite{Wang2018Visual} proposed a domain-invariant classifier MEDA in Grassmann manifold with structural risk minimization, while performing cross-domain distribution alignment of marginal and conditional distributions with different importances. Graph based manifold regularization $\mathcal{M}(\cdot)$ can be written as
\begin{equation}
\begin{split}
 & \mathcal{M}(X)=\sum_{i,j}W_{ij}(f(x_i)-f(x_j))^2=tr(F^T\mathcal{L}F)\\
\end{split}
\label{graph_fun}
\end{equation}
where $X$ is the data of source and target domain, $F$ is the predicted labels, $\mathcal{L}=D-W$ is the Laplacian matrix, $W_{ij}$ is the weight between sample $i$ and $j$, and $D$ is a diagonal matrix with $D_{ii}=\sum_i W_{ij}$. This term constrains the geometric structure preservation in label propagation and helps classifier adaptation. Although manifold regularizer can improve classifier adaptation performance, the fact is that the manifold assumption may not always hold, particularly when domain distribution does not match~\cite{Zhang2017MCTL}.

\subsection{Bayesian Classifier-Based}
In learning complex systems with limited data, Bayesian learning can well integrate prior knowledge to improve the weak generalization of models caused by data scarcity. For unsupervised domain adaptation, an underlying assumption in the kernel classifier and manifold classifier based models is that the conditional domain shift between domains can be minimized without relying on the target labels. Additionally, these methods are deterministic, which rely more on the expensive cross-validation for determining the underlying manifold space where the kernel mismatch between domains is effectively reduced. Recently, probabilistic model, i.e., Bayesian classifier based graphical models for DA/TL have been studied~\cite{Gonen2014Bayesian,Gonen2014Bayesian1,Liu2015Bayesian, Gholami2017Bayesian, Karbalayghareh2018Bayesian,Perrone2018Bayesian}, which aim to have better insights on the transferrable process from source domain to target domain.

In~\cite{Gonen2014Bayesian}, G$\ddot{o}$nen and Margolin firstly proposed graphical model, i.e., kernelized Bayesian transfer learning (KBTL) for domain adaptation. This work aims to seek a shared subspace and learn a coupled linear classifier in this subspace using a full Bayesian framework, solved by a variational approximation based inference algorithm. In~\cite{Gholami2017Bayesian}, Gholami et al. proposed a probabilistic latent variable model (PUnDA) for unsupervised domain adaptation, by simultaneously learning the classifier in a projected latent space and minimizing the MMD based domain disparity. A regularized Variational Bayesian (VB) algorithm was used for efficient model parameter estimation in PUnDA, because the computation of exact posterior distribution of the latent variables is intractable. More recently, Karbalayghareh et al.~\cite{Karbalayghareh2018Bayesian} proposed an optimal Bayesian transfer learning (OBTL) classifier to formulate the optimal Bayesian classifier (OBC) in target domain by using the prior knowledge of source and target domains, where OBC~\cite{Dalton2013OBC} aims to achieve Bayesian minimum mean squared error over uncertainty classes. In order to avoid costly computations such as MCMC sampling, OBTL classifier was derived based on the Laplace approximated hypergeometric functions.

\subsection{Discussion and Summary}
In this section, classifier adaptation including kernel classifier, manifold regularizer and Bayesian classifier are surveyed, which mostly rely on a small amount of tagged target domain data and facilitate semi-supervised transfer learning. This can be easily adapted to unsupervised transfer learning by pre-computing and iteratively updating the pseudo-labels of the completely unlabeled target domain in classifier adaptation. The kernel classifier focuses on SVM or MKL learning jointly with MMD based domain disparity minimization. The manifold regularizer based models aim to preserve the data affinity structure for label propagation. The Bayesian classifier based models resort to compensating the generalization performance loss due to data scarcity by modeling on the prior knowledge under reliable distribution assumptions, and having theoretical understanding on transfer learning from the viewpoint of data generation.

However, some inherent flaws exist: 1) incorrect pseudo-labels of target data significantly lead to performance degradation; 2) inaccurate distribution assumption in estimating various latent variables produces very negative effect; 3) the manifold assumption between domains does not hold for serious domain disparity.

\section{Deep Network Adaptation}
Deep neural networks (DNNs) have been recognized as dominant techniques for addressing computer vision tasks, due to their powerful feature representation and end-to-end training capability. Although DNNs can achieve more generalized features and performance in visual categorization, they rely on massive amounts of labeled data. For a target domain where the labeled data is unavailable or a very few labeled data is available, deep network adaptation started to rise. Yosinski et al.~\cite{Yosinski2014How} has discussed the transferability of features in bottom, middle and top layers of DNNs, and demonstrated that the transferability of features decreases as the distance between domains increases. In~\cite{Donahue2014DeCAF}, Donahue et al. proposed the deep convolutional activation feature (DeCAF) extracted by using a pre-trained AlexNet model~\cite{Alex2012Net}, which has well proved the generalization of DNNs for generic visual classification. This work further facilitated deep transfer learning and deep domain adaptation. Generally, the presented three types of TAL models in Section 3, 4 and 5, including instance re-weighting, feature adaptation and classifier adaptation, can be incorporated into DNNs with end-to-end training for deep network adaptation. In 2015, Long et al.~\cite{Long2015Learning,Long2018DAN} proposed a deep adaptation network (DAN) for learning transferrable features, which, for the first time opened the topic of deep transfer and adaptation. The basic idea of DAN is to enhance feature transferability in task-specific layers of DNNs by embedding the higher layered features into reproducing kernel Hilbert spaces (RKHSs) for nonparametric kernel matching (e.g., MMD-based) between domains. In training process, DAN was trained by fine-tuning on the ImageNet pre-trained DNN, such as AlexNet~\cite{Alex2012Net}, VGGNet~\cite{Simonyan2015VGGnet}, GoogLeNet~\cite{Szegedy2015GoogLeNet} and ResNet~\cite{He2015Resnet}.

Another sub-branch of deep network adaptation is partial domain adaptation~\cite{CaoZ2018cvpr,Zhang2018PDA,LiS2020TPAMI,CaoZJ2019cvpr} and open-set domain adaptation~\cite{PanY2020CVPR,Kundu2020CVPR,FangZ2020TNNLS}, which loose the category space (label set) consistency assumption across domains. The former supposes that the source domain shares partial categories with target data, i.e., $\mathcal{C}_S\supset\mathcal{C}_T$, and the disjoint source categories are recognized as noisy instances. The latter assumes that the target domain shares partial categories with source data, i.e., $\mathcal{C}_S\subset\mathcal{C}_T$, and the disjoint target categories are unified as \textit{unknown} category, i.e., the $(C+1)^{th}$ class. $C$ is the number of known classes. The common strategy is to actively discover the common/priviate classes with thresholding and re-weighting the model in sample-level. More general, You et al.~\cite{YouK2019cvpr} proposed a universal domain adaptation (UDA) which requires no prior knowledge on the label sets. That is, both domains can have private categories. Actually, the methodology in conventional DA can be freely generalized into PDA or OSDA with modifications, therefore, they are not further specified in order to put emphasis on the methodological aspects.

In recent years, various deep network adaptation models have emerged by modifying the network \textit{architecture}, \textit{loss} and \textit{threshold tricks}, which is presenting a blowout trend, because deep network adaptation has yield significant performance gains against shallow domain adaptation methods. For a better overview of various deep network adaptation methods, we divide them into (i) \textit{Marginal Alignment-Based}, (ii) \textit{Conditional Alignment-Based}, (iii) \textit{Batch Normalization Layer-Derived}, and (iv) \textit{Autoencoder-Based}, in which the first three mainly focus on convolutional neural network architecture. The taxonomy of deep network adaptation approaches is summarized in Table~\ref{tab4}.
\begin{table}[htbp]
\centering
\caption{Our Taxonomy of Deep Network Adaptation Approaches}
\begin{tabular}{cccccc}
 \hline
 \scshape \thead{Deep Net Adaptation} & \scshape Model Basis & \scshape Reference \\
 \hline
 \textbf{\thead{Marginal Alignment}} & CNN\&MMD & \thead{\cite{Long2015Learning,Tzeng2014Deep,Liu2016DDA},\\\cite{Long2017JAN,Long2016Unsupervised,Ge2017DTL}} \\
 \hline
 \textbf{Conditional Alignment} & \thead{CNN\&MMD\\\&Semantics} & \thead{\cite{Long2018DAN,Zhang2015DTN,Motiian2017udsd},\\\cite{ChenC2019CVPR,Deng2019ICCV,Cicek2019ICCV}\\\cite{Kang2019CVPR,WangS2020IJCAI}}\\
 \hline
 \textbf{BN Layer-Derived} & \thead{CNN\&Whitening} &
 \thead{\cite{Li2016arXiv,Carlucci2017ICCV,Roy2019CVPR},\\\cite{ChangC2019CVPR,WangX2019NIPS}} \\
 \hline
 \textbf{Autoencoder-Based} & \thead{Stacked Denoising\\autoencoders} & \thead{\cite{Glorot2011SDA,Chen2017DAE,Zhuang2015DAE},\\\cite{Zhou2014HHTL,Wen2019DTL,Ghifary2016Deep}}\\
 \hline
\end{tabular}
\label{tab4}
\end{table}
\subsection{Marginal Alignment-Based}
In unsupervised deep domain adaptation frameworks, for reducing the distribution disparity $d_{\mathcal{H}\Delta\mathcal{H}}(\mathcal{D}_S,\mathcal{D}_T)$ between labeled source domain and unlabeled target domain, the top layered features were generally transformed to a RKHS space where the maximum mean discrepancy (MMD) based kernel matching between domains was performed, which is recognized as marginal alignment based deep network adaptation~\cite{Long2015Learning,Tzeng2014Deep,Liu2016DDA,Long2017JAN}. For image classification, the softmax guided cross-entropy loss on the labeled source data is generally minimized. Representative works can be referred to as DDC proposed by Tzeng et al.~\cite{Tzeng2014Deep} and DAN~\cite{Long2015Learning}. The model can be written as
\begin{equation}
\begin{split}
 & \min_\Theta\frac{1}{N_s}\sum_{i=1}^{N_s}\mathcal{J}(\theta(x_i),y_i)
 +\lambda\sum_{l}d_{ma}^2(\mathcal{D}_s^l,\mathcal{D}_t^l)\\
\end{split}
\label{ma_fun}
\end{equation}
where $\mathcal{J(\cdot)}$ is the cross-entropy loss function, $\theta(\cdot)$ is the feature representation function, $\mathcal{D}^l$ denotes the domain feature set from the $l^{th}$ layer and $d_{ma}^2(\cdot)$ is the marginal alignment function (e.g., MMD in Eq.(\ref{MMD_fun})) between domains. Clearly, in Eq.(\ref{ma_fun}), multiple MMDs were formulated, one for each layer, and the summation of all MMDs is minimized. For better measuring the discrepancy between domains, a unified MMD called joint MMD (JMMD) was further designed by Long et al.~\cite{Long2017JAN} in a tensor product Hilbert space for matching the joint distribution of activations of multiple layers. Besides, other domain discrepancy metrics, such as Kullback-Leibler (KL) divergence~\cite{Pan2019CVPR,LeeS2019ICCV,TangH2020CVPR,LiuYenCheng2018CVPR}, Jensen-Shannon (JS) divergence~\cite{ZhaoH2019ICML}, Wasserstein distance~\cite{Lee2019CVPR,Balaji2019ICCV}, larger feature norms~\cite{XuR2019ICCV}, margin disparity discrepancy~\cite{ZhangY2019ICML}, moment distance~\cite{PengX2019ICCV}, and optimal transport distance~\cite{LiM2020CVPR, XuR2020CVPR} have also been explored.

The model in  Eq.(\ref{ma_fun}) does not take into account the network outputs of target domain stream, which may not well adapt the source classifier to target data. For addressing this problem, conditional-entropy minimization principle~\cite{Grandvalet2004em} that favors the low-density separation between classes in unlabeled target data $\mathcal{D}_t$ was further exploited in~\cite{Long2018DAN,Long2016Unsupervised,Ge2017DTL}. The entropy minimization is written as
\begin{equation}
\begin{split}
 & \min_{f\in\mathcal{F}}-\frac{1}{N_t}\sum_{i=1}^{N_t}\sum_{j=1}^Cf_j(x_i^t)\log f_j(x_i^t)\\
\end{split}
\label{em_fun}
\end{equation}
where $f_j(x)$ is the probability that sample $x$ is predicted as class $j$. Entropy minimization is amount to \textit{uncertainty} minimization of the predicted labels of target samples. Additionally, by following the assumption of ASVM in~\cite{YangYanHauptmann2007}, the residual $\Delta f(x)$ between source and target classifiers was learned in the residual transfer network (RTN)~\cite{Long2016Unsupervised}, with a residual connection.

\subsection{Conditional Alignment-Based}
In marginal alignment based deep network adaptation, only the top layered feature matching in RKSH spaces was formulated by using the nonparametric MMD metric. However, the high-level semantic information was not taken into account in domain alignment, which may degrade the adaptability of source data trained DNNs to unlabeled target domain due to the class mis-alignment. Theoretically, an unexpected larger upper bound of the joint error $\lambda=\min_{h\in\mathcal{H}}\epsilon_S(h,l_s)+\epsilon_T(h,l_t)\leq \min_{h\in\mathcal{H}}\epsilon_S(h,l_s)+\epsilon_T(h,l_s)+\epsilon_T(l_s,l_t)$ may be encountered, where $l_s$ and $l_t$ are labeling function for source and target domains, respectively. This is because $\epsilon_T(l_s,l_t)$ can be large enough due to class mis-alignment. Therefore, conditional alignment based deep network adaptation methods were presented jointly with marginal alignment based models~\cite{Zhang2015DTN,Motiian2017udsd,Kang2019CVPR,ChenC2019CVPR,Deng2019ICCV,Cicek2019ICCV,WangS2020IJCAI}. Due to the unavailability of target labels, in such kinds of models, clustering labels or progressively updated target pseudo-labels (label refinement) are generally exploited. Similar to the formulation of MMD in Eq.(\ref{MMD_fun}), the conditional alignment was generally formulated by building between-domain MMD like discrepancy metric $d_{ca}^2$ on the probability $p$, which reflects the uncertainty predicting a sample to class $c$.
\begin{equation}
\begin{split}
 & d_{ca}^2=\sum_{c=1}^C \Arrowvert \frac{1}{n_s}\sum_{i=1}^{n_s}p(y_i^s=c\vert x_i^s)-\frac{1}{n_t}\sum_{j=1}^{n_t}p(y_i^t=c\vert x_j^t)\Arrowvert^2\\
\end{split}
\label{ca_fun}
\end{equation}
Therefore, conditional alignment based deep adaptation model was generally constructed by combining Eq.(\ref{ma_fun}) and Eq.(\ref{ca_fun}) together. The probability constraint between domains can effectively improve the semantic discrimination. Actually, $l_1$-norm can also be imposed on the difference between the probabilities of source and target samples. Generally, the basic strategies for constraining the joint error $\lambda$ to be small enough refer to as target pseudo-labeling~\cite{ChenC2019CVPR}, source pre-trained network freezing~\cite{LiuH2019ICML}, information-theory metric~\cite{ZhaoH2019ICML}, etc.

Very recently, with the privacy protection mechanism, \textit{source-free} domain adaptation models have emerged~\cite{LiR2020cvpr,KunduJN2020cvpr}, in which the source samples are used to pre-train a CNN model but not explicitly exploited in DA model.

\subsection{Batch Normalization Layer-Derived}
Other than loss functions design for TAL, in this section, we introduce another branch, i.e., network design. That is modifying the internal component of CNN for transferable representation. Batch normalization (BN) has been proved to improve the predictive performance and training efficiency by normalizing the inputs to be zero-mean and univariance, and further scales and shifts the normalized signals with two trainable parameters.Given a batch $B$, each $\mathbf{x}_i\in B$ is transformed as $BN(x_{i,k})=\gamma_k\frac{x_{i,k}-\mu_{B,k}}{\sqrt{\sigma^2_{B,k}+\epsilon}}+\beta_k$, where $k~(1\leq k\leq d)$ indicates the $k$-th dimension of $\mathbf{x}_i$, $\mu_{B,k}$ and $\sigma_{B,k}$ are the mean and std. w.r.t. the $k$-th dimension in batch $B$, $\epsilon$ is a constant for numerical stability, $\gamma_k$ and $\beta_k$ are learnable scaling and shifting parameters. For cross-domain applications, due to the datasets bias and domain shift, the means and variance in a batch for different domains should be different. Therefore, researchers have proposed to embed domain alignment layers into the network based on batch normalization layer for domain adaptation~\cite{Li2016arXiv,Carlucci2017ICCV,Roy2019CVPR,ChangC2019CVPR,WangX2019NIPS}.

Roy et al.~\cite{Roy2019CVPR} proposed a domain specific batch whitening (BW) layer instead of BN for reducing domain shift, which is defined as $BW(x_{i,k})=\gamma_k\hat{x}_{i,k}+\beta_k$ and $\hat{\mathbf{x}}_i=W_B(\mathbf{x}_i-\mathbf{\mu}_B)$, such that $W^T_{B}W_B=\Sigma_B^{-1}$, where $\mu_B$ is the domain-specific mean vector in a batch $B$ and $\Sigma_B^{-1}$ is the domain-specific covariance matrix computed using $B$. Wang et al.~\cite{WangX2019NIPS} proposed a transferable normalization (TransNorm), in which domain specific mean vectors $\mu_S, \mu_T$ and variances $\sigma_S, \sigma_T$ are computed, while the learnable parameters $\gamma$, $\beta$ to scale and shift the normalized signal are domain shared. Similarly, the AutoDial proposed by Carlucci et al.~\cite{Carlucci2017ICCV} also shares the two parameters across domains and Chang et al.~\cite{ChangC2019CVPR} also proposed a domain specific batch normalization layer.
Besides BN layer, it is worthy noting that, Lee~\cite{LeeS2019ICCV} proposed a dropout adaptation (DTA) by imposing a learnable dropout mask.

\subsection{Autoencoder-Based}
As mentioned above, the training of DNNs needs a large amount of labeled source data. For unsupervised feature learning in domain adaptation, deep autoencoder based network adaptation framework was presented~\cite{Glorot2011SDA,Chen2017DAE,Zhuang2015DAE,Zhou2014HHTL,Wen2019DTL,PengX2019ICML}. Generic auto-encoders are comprised of an encoder function $f(\cdot)$ and a decoder function $g(\cdot)$, which are typically trained to minimize the reconstruction error. Denoising autoencoders (DAE) were generally constructed with one-layer neural networks for reconstructing original data from partially or randomly corrupted data~\cite{Vincent2008DAE}. The denoising autoencoders can be stacked into a deep network (i.e., SDA), optimized by greedy layer-wise fashion based on stochastic gradient descent (SGD). The rational behind deep autoencoder based network adaptation is that the source data trained parameters of encoder and decoder can be adapted to represent those samples from a target domain.

In~\cite{Glorot2011SDA}, Glorot et al. proposed a SDA based feature representation in conjunction with SVMs for sentiment analysis across different domains.
Chen et al.~\cite{Chen2017DAE} proposed a marginalized stacked denoising autoencoder (mSDA), which addressed two crucial limitations of SDAs, such as high computational cost and low scalability to high-dimensional features, by inducing a closed-form solution of parameters without SGD. In~\cite{Zhuang2015DAE}, Zhuang et al. proposed a supervised deep autoencoder for learning domain invariant features. The encoder is constructed with two encoding layers: embedding layer for domain disparity minimization and label encoding layer for softmax guided source classifier training. Suppose $x$, $z$ and $\hat{x}$ to be the input sample, intermediate representation (encoded) and reconstructed output (decoded), respectively, then there is
\begin{equation}
\begin{split}
 & z=f(x), \hat{x}=g(z)
\end{split}
\label{en_de_fun}
\end{equation}
where $z$ is the intermediate feature representation of sample $x$. Generally, stacked deep autoencoder based TAL framework can be written as
\begin{equation}
\begin{split}
 & \min_{f,g,\theta}\mathcal{J}(x,\hat{x})+\lambda\Omega(z_s,z_t)+\beta\mathcal{L}(z_s,y_s,\theta)
 +\gamma\mathcal{R}(f,g)\\
\end{split}
\label{encoder_fun}
\end{equation}
where $f$ is domain shared encoder, $g$ is domain shared decoder, $\mathcal{J}(\cdot)=\mathcal{J}_s(\cdot)+\mathcal{J}_t(\cdot)$ represents the reconstruction error loss (e.g., $l_2$-norm squared loss), $\Omega(\cdot)$ is the distribution discrepancy metric between source feature $z_s$ and target feature $z_t$, $\mathcal{L}(\cdot)$ is the classifier loss (e.g. cross-entropy) with parameter $\theta$ learned on the set$(z_s,y_s)$, and $\mathcal{R}(\cdot)$ is the regularizer of the network parameters of $f$ and $g$. In~\cite{Zhuang2015DAE}, Kullback-Leibler (KL) divergence~\cite{Kullback1987} based distribution distance metric was considered. KL is a non-symmetric measure of the divergence between two probability distributions $P$ and $Q$, which was defined as $D_{KL}(P\vert\vert Q)=\sum_{i}P(i)\ln(\frac{P(i)}{Q(i)})$. Smaller value of $D_{KL}(\cdot)$ means higher similarity of two distributions. Due to that $D_{KL}(P\vert\vert Q)\neq D_{KL}(Q\vert\vert P)$, a symmetric KL version was used in~\cite{Zhuang2015DAE}, in which the $\Omega(\cdot)$ in Eq.(\ref{encoder_fun}) was written as
\begin{equation}
\begin{split}
 &\Omega(z_s,z_t)=D_{KL}(P_s\vert\vert P_t)+D_{KL}(P_t\vert\vert P_s)\\
\end{split}
\label{kl_fun}
\end{equation}
where $P_s=\frac{\bar{z}_s}{\Sigma\bar{z}_s}$ and $P_t=\frac{\bar{z}_t}{\Sigma\bar{z}_t}$ represent the probability distribution of source and target domains. $\bar{z}_s$ and $\bar{z}_t$ represent the mean vector of encoded feature representations of source and target samples, respectively.

Similar to the reconstruction protocol in stacked autoencoder, a related work with deep reconstruction based on convolutional neural networks can be referred to as~\cite{Ghifary2016Deep}, in which the encoded source feature representation is feeded into the source classifier for visual classification and simultaneously into the decoder module for reconstructing the target data. Under this framework, a shared encoder for both domains can be learned.

\subsection{Discussion and Summary}
In this section, deep network adaptation advances are presented and categorized, which mainly contains three types of technical challenges: marginal alignment based, conditional alignment based and autoencoder based. A common characteristic of these methods is that the softmax guided cross-entropy loss based on labeled source data was minimized for classifier learning. In marginal alignment based models, the distribution discrepancy of feature representation from top layers is generally characterized by MMD. Besides that, the semantic similarity across domains was further characterized in conditional alignment based models. Different from both marginal and conditional alignment models, the autoencoder based ones tend to learn domain invariant feature embedding by imposing a Kullback-Leibler divergence in feature embedding layer.

Despite recent advances deep network adaptation faces several challenges: 1) a number of labeled source data is needed for training (fine-tuning) a deep network; 2) the confidence of an unlabeled target sample predicted to class $k$ is sometimes very low when domain disparity is very large; 3) the interpretability of deep adaptation is not optimistic and negative transfer is easily encountered.

\section{Adversarial Adaptation}
Adversarial learning, originated from the generative adversarial net (GAN)~\cite{Goodfellow2014GAN}, is a promising approach for generating pixel-level target samples or feature-level target representations by training robust DNNs. Currently, adversarial learning has become an increasing popular idea for addressing TAL issues, by minimizing the between-domain discrepancy through an adversarial objective (e.g., binary domain discriminator), instead of the generic MMD-based domain disparity in RKHS spaces. In fact, minimization of the domain disparity is amount to domain confusion in a learned feature space, where the domain discriminator cannot discriminate which domain a sample comes from. In this paper, the adversarial adaptation based TAL approaches are divided into three types: (i) \textit{Gradient Reversal-Based}, (ii) \textit{Minimax Optimization-Based} and (iii) \textit{Generative Adversarial Net-Based}. The first two resort to feature-level domain confusion supervised by a domain discriminator for domain distribution discrepancy minimization, while the last one tends to pixel-level domain transfer by synthesizing implausible target domain images. Actually, domain adaptation is in essence a minimax problem. The taxonomy of adversarial adaptation approaches is summarized in Table~\ref{tab5}.

\begin{table}[htbp]
\centering
\caption{Our Taxonomy of Adversarial Adaptation Approaches}
\begin{tabular}{cccccc}
 \hline
 \scshape \thead{Adversarial Adaptation} & \scshape Model Basis & \scshape Reference \\
 \hline
 \textbf{\thead{Gradient Reversal}} & \thead{Domain\\ Confusion\\\&GRL} & \thead{\cite{Ganin2015Unsupervised,Pinheiro2018sl,Zhang2018PDA},\\\cite{Pei2018MADA,Zhang2018CAN,Chen2018RCNN},\\\cite{Duan2017advnet,Duan2017advkin,Dubey2018PC}} \\
 \hline
 \textbf{\thead{Minimax Optimization}} & \thead{Domain\\Confusion\\\&Game} & \thead{\cite{Tzeng2015Simultaneous,Ajakan2015DANN,Bousmalis2016Domain}\\\cite{E.Tzeng2017dversarial,Motiian2017ADA,Rozantsev2018RPT},\\\cite{Mingsheng2018Conditional,Xie2018MSTN,Saito2018MCD}} \\
 \hline
 \textbf{GANs-Based} & \thead{Pixel-level\\Image Synthesis} & \thead{\cite{Hoffman2017CyCADA,Bousmalis2017PLDA,Taigman2017CDIG},\\\cite{Hu2018DGAN,Murez2018IIT,Choi2018Star},\\\cite{Hong2018CGAN,Wei2018PTGAN,Zhong2018CSA}\\\cite{Zheng2017ReIDGAN,Yin2017Towards,Hu2018pose}}\\
 \hline
\end{tabular}
\label{tab5}
\end{table}

\subsection{Gradient Reversal-Based}
In adversarial optimization of DNNs between the general cross-entropy loss for source classifier learning and the domain discriminator for domain label prediction, Ganin and Lempitsky~\cite{Ganin2015Unsupervised} firstly demonstrated that the domain adaptation behavior can be achieved by adding a simple but effective \textit{gradient reversal layer (GRL)}. The augmented deep architecture can still be trained using standard stochastic gradient descent (SGD) based backpropagation. The gradient reversal based adversarial adaptation network consists of three parts: domain-invariant feature representation $\theta_f$, visual classifier $\theta_c$ and domain classifier $\theta_d$. Objectively, $\theta_f$ can be learned by trying to \textit{minimize} the visual classifier loss $L_c$ and simultaneously \textit{maximize} the domain classifier loss $L_d$, such that the feature representation can be domain invariant (i.e. domain confusion) and class discriminative. Therefore, in backpropagation optimization of $\theta_f$, the contributed gradients from losses $L_c$ and $L_d$ are $\frac{\partial{L_c}}{\partial{\theta_f}}$ and $-\lambda\frac{\partial{L_d}}{\partial{\theta_f}}$, respectively. The essence of GRL lies in the reversal gradient with negative multiplier $-\lambda I$.

More recently, the gradient reversal based adversarial strategy has been used for domain adaptation~\cite{Pinheiro2018sl,Zhang2018PDA,Pei2018MADA,Zhang2018CAN,LiS2019MM} under CNN architecture, domain adaptive object detection~\cite{Chen2018RCNN,HeZ2019iccv,HeZ2020eccv} under Faster-RCNN framework, large-scale kinship verification~\cite{Duan2017advnet,Duan2017advkin} and fine-grained visual classification~\cite{Dubey2018PC} under Siamese network. By following a similar protocol with~\cite{Ganin2015Unsupervised}, in~\cite{Pinheiro2018sl,Chen2018RCNN}, a binary domain classifier (e.g., source is labeled as +1 and -1 for target) was designed as an adversarial objective for learning domain-invariant features by deploying a GRL layer. In~\cite{Duan2017advnet,Duan2017advkin}, two methods, AdvNet and Adv-Kin were proposed, in which a general Siamese network was constructed with three fully-connected (\textit{fc-}) layers for similarity learning. The reversal gradient with negative multiplier $-\lambda I$ was placed in the $1^{st}$ \textit{fc}-layer (MMD-loss), the generic contrastive loss was deployed in the $2^{nd}$ \textit{fc}-layer and the softmax guided cross-entropy loss was deployed in the last \textit{fc}-layer. In~\cite{Pei2018MADA}, Pei et al. argued that single domain discriminator based adversarial adaptation only aligns the between-domain distribution without exploiting the multimode structures. Therefore, they proposed a multi-adversarial domain adaptation (MADA) method based on GRL with multiple class-wise domain discriminators for capturing multimode structures, such that fine-grained alignment of different distributions is enabled. Also, Zhang et al.~\cite{Zhang2018CAN} proposed a collaborative adversarial network (CAN) by designing multiple domain classifiers, one for each feature extraction block in CNN.

\subsection{Minimax Optimization-Based}
In GANs, the two key parts $G$ and $D$ are often placed with an adversarial state, and generally solved by using a minimax based gaming optimization method~\cite{Goodfellow2014GAN}. Therefore, the minimax optimization based adversarial adaptation can be implemented for domain confusion, through an adversarial objective of the domain discriminator or regressor~\cite{Tzeng2015Simultaneous,Ajakan2015DANN,Bousmalis2016Domain,E.Tzeng2017dversarial,Motiian2017ADA,Rozantsev2018RPT,Mingsheng2018Conditional,Xie2018MSTN,WangS2020IJCAI}. Minimax optimization based adversarial adaptation training of DNNs originated in 2015~\cite{Tzeng2015Simultaneous,Ajakan2015DANN}. Domain confusion maximization based adversarial domain adaptation was first proposed by Tzeng et al.~\cite{Tzeng2015Simultaneous}, in which an adversarial CNN framework was deployed with classification loss, soft label loss and two adversarial objectives i.e., domain confusion loss and domain discriminator loss. In~\cite{Ajakan2015DANN}, Ajakan et al. firstly proposed an adversarial training of stacked autoencoders (DANN) deployed with classification loss and an adversarial objective i.e., domain regressor loss.

Suppose the labeled source data trained visual classifier to be $C$, the domain discriminator to be $D$, and the feature representation to be $F$. The corresponding parameters are defined as $\theta_C$, $\theta_D$ and $\theta_F$. The general adversarial adaptation model aims to minimize the visual classifier loss $\mathcal{L}_C$ and maximize the domain discriminator loss $\mathcal{L}_D$ by learning $\theta_F$, such that the feature representation function $F$ can be more discriminative and domain-invariant. Simultaneously, the adversarial training aims to minimize the domain discriminator loss $\mathcal{L}_D$ under $\theta_F$. Generally, maximizing $\mathcal{L}_D$ is amount to maximizing the domain confusion, such that it cannot discriminate which domain the samples come from, and vice versa. The above process can be generally formulated as the following adversarial adaptation model,
\begin{equation}
\begin{split}
 &\min_{\theta_C,\theta_F} \mathcal{L}_C(\mathcal{D}_S, \mathcal{Y}_S; \theta_C,\theta_F)-\lambda\mathcal{L}_D(\mathcal{D}_S,\mathcal{D}_T,\theta_D; \theta_F)\\
 &\qquad\qquad\qquad \min_{\theta_D} \mathcal{L}_D(\mathcal{D}_S,\mathcal{D}_T,\theta_F;\theta_D)\\
\end{split}
\label{adversarial_fun}
\end{equation}
where $\mathcal{D}_S$ and $\mathcal{D}_T$ mean the source and target domain samples, and $\mathcal{Y}_S$ denotes the source data labels.

Under this basic framework in Eq.(\ref{adversarial_fun}), Tzeng et al.~\cite{E.Tzeng2017dversarial} further proposed an adversarial discriminative domain adaptation (ADDA) method, in which two CNNs were separately learned for source and target domain. The training of source CNN relied only on the source data and labels by minimizing the cross-entropy loss $\mathcal{L}_C$, while the target CNN and the domain discriminator loss $\mathcal{L}_D$ was alternatively trained in an adversarial fashion with the source CNN fixed. Rozantsev et al.~\cite{Rozantsev2018RPT} proposed a residual parameter transfer model with adversarial domain confusion supervised by a domain classifier, in which the residual transform between domains was deployed in convolutional layers. For augmenting the domain-specific feature representation, Long et al.~\cite{Mingsheng2018Conditional} proposed a conditional domain adversarial network (CDAN), in which the feature representation and classifier prediction were integrated via multilinear map for jointly learning the domain classifier. More recently, Saito et al.~\cite{Saito2018MCD} proposed a novel adversarial strategy, i.e., maximum classifier discrepancy (MCD), which aims to maximize the discrepancy between two classifiers' outputs instead of domain discriminator. The feature extractor aims to minimize the two classifiers' discrepancy. They argued that the general domain discriminator does not take into account the task-specific decision boundaries between classes, which may lead to ambiguous features near class boundaries from the feature extractor. In MCD, only two classifiers are designed. By following \textit{the wisdom of crowd} principle, a stochastic classifier (STAR)~\cite{LuZ2020CVPR} which can sample an arbitrary number of classifiers from a Gaussian distribution rather than two classifiers is proposed and achieves better performance. By following the adversarial principle of double classifiers, a classifier competition model for reliable domain adaptation was proposed in~\cite{FuJ2020TCASII}.

\subsection{Generative Adversarial Net-Based}
In generative adversarial net (GAN)~\cite{Goodfellow2014GAN} and its variants, two key parts: generator $G$ and discriminator $D$ are generally composed. The generator $G$ aims to synthesize implausible images by using the encoder and decoder, while the discriminator $D$ plays a role in identification of authenticity by recognizing a sample to be true or false. A minimax gaming based alternative optimization scheme is generally used for solving $G$ and $D$. In TAL studies, started from 2017, GAN based models have been presented to synthesize distribution approximated pixel-level images with target domain and then enable the cross-domain image classification by using synthesized image samples (e.g., objects, scenes, pedestrians and faces, etc.)~\cite{Hoffman2017CyCADA,Bousmalis2017PLDA,Taigman2017CDIG, Hu2018DGAN,Murez2018IIT,Choi2018Star,Hong2018CGAN,Wei2018PTGAN,Zhong2018CSA,Sankaranarayanan2018CVPR,XuRuijia2018CVPR}.

Under the CycleGAN framework proposed by Zhu et al.~\cite{Zhu2017Unpaired}, Hoffman et al.~\cite{Hoffman2017CyCADA} firstly proposed a cycle-consistent adversarial domain adaptation model (CyCADA) for adapting representations in both pixel-level and feature-level without requiring aligned pairs, by jointly minimizing pixel loss, feature loss, semantic loss and cycle consistence loss.
Bousmalis et al.~\cite{Bousmalis2017PLDA} and Taigman et al.~\cite{Taigman2017CDIG} proposed GAN-based models for unsupervised image-level domain adaptation, which aims to adapt source domain images to appear as if drawn from target domain with well-preserved identity. In~\cite{Hu2018DGAN}, Hu et al. proposed a duplex GAN (DupGAN) for image-level domain transformation, in which the duplex discriminators, one for each domain, were trained against the generator for ensuring the reality of the domain transformation. Murez et al.~\cite{Murez2018IIT} and Hong et al.~\cite{Hong2018CGAN} proposed image-to-image translation based domain adaptation models by leveraging GAN and synthetic data for semantic segmentation of the target domain images. Person re-identification (ReID) is typically a cross-domain feature match and retrieval problem~\cite{Li2015CPDL,Chen2018ReID}. Recently, for addressing ReID challenges in complex scenarios, GAN-based domain adaptation was presented for implausible person image generation from source to target domain~\cite{Zheng2017ReIDGAN, Wei2018PTGAN,Zhong2018CSA}, across different visual cues and styles, such as poses, backgrounds, lightings, resolutions, seasons, etc. Additionally, GAN based cross-domain facial image generation for pose-invariant face representation, face frontalization and rotation were intensively studied~\cite{Tran2017DRGAN,Huang2017Beyond,Yin2017Towards,Hu2018pose}, all of which tend to address domain adaptation and transfer problems in face recognition across poses.

\subsection{Discussion and Summary}
In this section, adversarial adaptation is presented with three streams, including gradient reversal, minimax optimization and generative adversarial net (GAN). The gradient reversal and minimax optimization share a common characteristic, i.e., feature-level adaptation, by introducing a domain discriminator based adversarial objective for training against the feature extractor. The difference between them is the \textit{against strategy}. Different from both of them, GAN-based adversarial adaptation focuses on pixel-level adaptation, i.e., image generation from source domain to a target, such that the synthesized implausible images are as if drawn from target domain.

Adversarial adaptation is recognized to be an emerging perspective, despite these advances it still faces with several challenges: 1) the domain discriminator is easily overtrained; 2) maximizing only domain confusion easily leads to class bias; 3) the gaming between feature generator and discriminator is human dependent.
\section{Benchmark Datasets of VTAL}
In this section, the benchmark datasets for testing TAL models are introduced to facilitate readers' impression on how to start studies of transfer adaptation learning. Totally, 12 benchmark datasets including Office-31 (3DA)~\cite{Saenko2010symm}, Office+Caltech-10 (4DA)~\cite{Saenko2010symm,GongShiShaEtAl2012,Donahue2014DeCAF,Griffin2007Caltech}, MNIST+USPS~\cite{Zhang2016LSDT,XuFangWuEtAl2016}, Multi-PIE~\cite{Zhang2016LSDT,XuFangWuEtAl2016}, COIL-20~\cite{Rate2011Columbia}, MSRC+VOC2007~\cite{LongWangDingEtAl2014}, IVLSC~\cite{Li2017Deeper,Ghifary2015Domain}, AwA~\cite{C.H.Lampert2009Learning}, Cross-dataset Testbed~\cite{TorralbaEfros2011}, Office Home~\cite{Venkateswara2017Deep}, ImageCLEF~\cite{Caputo2014ImageCLEF}, and P-A-C-S~\cite{Li2017Deeper} are summarized, each of which contains at least 2 different domains. For these benchmarks, classification accuracy of target data is commonly used for performance comparison. Due to the space limitation, the classification performances of different models are not listed, so that we can focus more on discussing the methodological advances and potential issues.

\subsection{Office-31 (3DA)}
Office-31 is a popular benchmark for visual domain transfer, which includes 31 categories of samples drawn from three different domains, i.e., Amazon (A), DSLR (D) and Webcam (W). Amazon consists of online e-commerce pictures, DSLR contains high-resolution pictures and Webcam contains low-resolution pictures taken by a web camera. There are totally 4652 images, composed of 2817, 498 and 795 images from domain A, D and W, respectively. In feature extraction, (1) for shallow features, 800-dimensional feature vectors extracted by the Speed Up Robust Features (SURF) were used, and (2) for deep features, 4096-dimensional feature vectors extracted from pre-trained AlexNet/VGG-net/ResNet-50 were generally used. In model evaluation, six kinds of source-target domain pairs were tested, i.e., A$\rightarrow$D, A$\rightarrow$W, D$\rightarrow$A, D$\rightarrow$W , W$\rightarrow$A, W$\rightarrow$D.

\subsection{Office+Caltech-10 (4DA)}
This 4DA dataset contains 4 domains, in which 3 domains (A, D, W) are from the Office-31 and another domain (C) is from Caltech-256, a benchmark containing 30,607 images of 256 classes in object recognition. The common 10 classes among the Office-31 and Caltech-256 were selected to form the 4DA, and therefore 2,533 images composed of 958, 157, 295 and 1123 images from domain A, D, W and C were collected. In evaluation, 12 tasks with different source-target domain pairs are addressed, i.e., A$\rightarrow$D, A$\rightarrow$C, A$\rightarrow$W, D$\rightarrow$A, D$\rightarrow$C, D$\rightarrow$W, C$\rightarrow$A, C$\rightarrow$D, C$\rightarrow$W, W$\rightarrow$A, W$\rightarrow$C, W$\rightarrow$D.

\subsection{MNIST+USPS}
MNIST and USPS are two benchmarks containing 10 categories of digit images under different distribution for handwritten digit recognition, and therefore qualified for TAL tasks. The MNIST includes 60,000 training pictures and 10,000 test pictures. The USPS includes 7291 training pictures and 2007 test pictures. For TAL tasks, 2000 pictures and 1800 pictures were randomly selected from MNIST and USPS, respectively. For feature extraction, each image was resized into 16$\times$16 and a 256-dimensional feature vector that encode the pixel values was finally extracted. In evaluation, 2 cross-domain tasks, i.e., MNIST$\rightarrow$USPS and USPS$\rightarrow$MNIST are addressed.

\subsection{Multi-PIE}
Multi-PIE is a benchmark with poses, illuminations and expressions in face recognition, which includes 41,368 faces of 68 different identities. For TAL tasks, (1) face recognition across poses is generally evaluated on five different face orientations, including C05: left pose, C07: upward pose, C09: downward pose, C27: front pose and C29: right pose. Totally, 3332, 1629, 1632, 3329, and 1632 facial images are contained in C05, C07, C09, C27 and C29. Therefore, 20 tasks were evaluated, i.e., C05$\rightarrow$C07, C05$\rightarrow$C09, C05$\rightarrow$C27, etc.; (2) face recognition across illuminations and exposure conditions is evaluated by randomly selecting two sets: PIE1 and PIE2 from front face images. Two tasks: PIE$\rightarrow$PIE2 and PIE2$\rightarrow$PIE1 were evaluated.

\subsection{COIL-20}
COIL-20 is a 3D object recognition benchmark containing 1440 images of 20 object categories. By rotating each object class horizontally of 5 degrees, 72 images per class after rotating 360 degrees were obtained. For TAL tasks, two disjoint subsets with different distribution i.e., COIL1 and COIL2 were prepared, where COIL1 contains the images in [$0^{\circ}$, $85^{\circ}$] U [$180^{\circ}$, $265^{\circ}$] and the images of COIL2 are in [$90^{\circ}$, $175^{\circ}$] U [$270^{\circ}$, $355^{\circ}$]. Therefore, two cross-domain tasks i.e., COIL1$\rightarrow$COIL2 and COIL2$\rightarrow$COIL1 were evaluated.

\subsection{MSRC+VOC2007}
The MSRC contains 4323 images of 18 categories and VOC2007 contains 5011 images of 20 categories. 1269 and 1530 images w.r.t six common categories, i.e., \textit{aeroplane, bicycle, bird, car, cow} and \textit{sheep}, were finally selected from MSRC and VOC2007, respectively. In feature representation, 128-dimensional DenseSIFT features were extracted for cross-domain image classification tasks, i.e., MSRC$\rightarrow$VOC2007 and VOC2007$\rightarrow$MSRC.

\subsection{IVLSC}
IVLSC is a large-scale image dataset containing five subsets, i.e., ImageNet (I), VOC2007 (V), LabelMe (L), SUN09 (S), and Caltech (C). For TAL tasks, 7341, 3376, 2656, 3282, and 1415 samples w.r.t. five common categories i.e., \textit{bird, cat, chair, dog} and \textit{human}, were randomly selected from I, V, L, S, and C domains, respectively. In feature representation, 4096-dimensional DeCaf6 deep features were extracted for cross-domain image classification under 20 tasks, i.e., I$\rightarrow$V, I$\rightarrow$L, I$\rightarrow$S, I$\rightarrow$C, ..., C$\rightarrow$I, C$\rightarrow$V, C$\rightarrow$L, C$\rightarrow$S.

\subsection{AwA}
AwA is an animal identification dataset containing 30,475 images of 50 categories, which provides a benchmark due to the inherent data distribution difference. This data set is currently less used in evaluating TAL algorithms.

\subsection{Cross-dataset Testbed}
This benchmark contains 10,473 images of 40 categories, collected from three domains: 3,847 images in Caltech256 (C), 4,000 images in ImageNet (I), and 2,626 images in SUN (S). In feature extraction, the 4096-dimensional DeCAF7 deep features were used for cross-domain image classification tasks, i.e., C$\rightarrow$I, C$\rightarrow$S, I$\rightarrow$C, I$\rightarrow$S, S$\rightarrow$C, S$\rightarrow$I.

\subsection{Office Home}
Office Home is a relatively new benchmark containing 15,585 images of 65 categories, collected from 4 domains, i.e., (1) Art (Ar): artistic depictions of objects in the form of sketches, paintings, ornamentation, etc.; (2) Clipart (Cl): collection of clipart images; (3) Product (Pr): images of objects without a background, akin to the Amazon category in Office dataset; (4) Real-World (RW): images of objects captured with a regular camera. In detail, there contains 2421, 4379, 4428 and 4357 images in \textit{Ar, Cl, Pr} and \textit{RW} domains, respectively. In evaluation, 12 cross-domain tasks were tested, e.g., Ar$\rightarrow$Cl, Ar$\rightarrow$Pr, Ar$\rightarrow$RW, Cl$\rightarrow$Ar, etc.

\subsection{ImageCLEF}
This benchmark includes 1800 images of 12 categories, which were drawn from 3 domains: 600 images in Caltech 256 (C), 600 images in ImageNet ILSVRC2012 (I), and 600 images in Pascal VOC2012 (P). Therefore, 6 cross-domain tasks i.e., C$\rightarrow$I, C$\rightarrow$P, I$\rightarrow$C, I$\rightarrow$P, P$\rightarrow$C, P$\rightarrow$I were evaluated.

\subsection{P-A-C-S}
PACS is a new benchmark containing 7 common categories: \textit{dog, elephant, giraffe, guitar, horse, house} and \textit{person}, from 4 domains, i.e., 1670 images in Photo (P), 2048 images in Art Painting (A), 2344 images in Cartoon (C), and 3929 images in Sketch (S). In feature representation, 4096-dimensional VGG-M deep features were used and 12 cross-domain tasks are evaluated, e.g., P$\rightarrow$A, P $\rightarrow$C, P$\rightarrow$S, A$\rightarrow$P, A$\rightarrow$C, etc.

\subsection{Discussion and Summary}
In this section, 12 benchmarks constructed based on popular datasets in computer vision such as ImageNet, ILSVRC, PASCAL VOC, Caltech-256, multi-PIE and MNIST for addressing cross-domain image classification tasks and evaluating the DA models are presented. Despite these endeavors made by researchers, more benchmarks in cross-domain vision understanding problems we could see, namely: object detection, semantic segmentation, visual relation modeling, scene parsing, etc. are future challenges for more universal and safe applications. These can better testify the \textit{practicality} of TAL methodologies.

Pleasantly, with the stronger models and algorithms, the performance on the benchmarks has taken a big step forward. However, we have to denote that the existing UDA setting has always used target data for training, which, inevitably leads to overfitting and loses fairness. Therefore, we must be aware of this, and a reasonable and scientific training and testing protocol is expected to effectively evaluate the DA models. Considering the necessity of target data, a possible solution is that, as few-shot learning does, partial unlabeled target samples are used as seen samples (training) for determining an ideal hypothesis $h^*$ with small joint error $\lambda=\epsilon_S(h^*)+\epsilon_T(h^*)$, and the unseen target samples (test) are used for model evaluation.

\section{VTAL in More Vision Tasks}
The first 8 sections presents the preliminaries, methodological advances and benchmarks of VTAL for image classification, a fundamental vision task. In this section, from the perspective of applications, we have a discussion on VTAL theory utilized in more vision tasks, such as object detection and semantic segmentation, as illustrated in Fig.~\ref{fig1}. It is not difficult to imagine that, for example, fog caused image degradation will inevitably lead to detection performance degradation. A preliminary research in~\cite{PeiYT2019TPAMI} finds that degradation removal does not help CNN-based image classification, because degradation removal focuses on pixel manipulation for image enhancement but does not bring any new information beneficial to classification. Therefore, cross-dataset object detection/segmentation (e.g., from cityscapes to foggy cityscapes) rather than learning to dehaze is an important but challenging vision task. In the past two years, this topic has attracted researcher' attention and a significant progress is witnessed. Specifically, many domain adaptive object detection models~\cite{Chen2018RCNN,HeZ2019iccv,SaitoK2019cvpr,HeZ2020eccv,ZhuXG2019cvpr,XuCD2020cvpr,ChenCQ2020cvpr,ZhengY2020cvpr} and domain adaptive semantic segmentation models~\cite{VuTH2019cvpr,LuoYW2019cvpr,LiYS2019cvpr,ChenYC2019cvpr,VuTH2019iccv,ChenMH2019iccv,LvFM2020cvpr,ZhangYH2020cvpr} have been proposed. These works have well revealed that the versatile VTAL theory and algorithm can be combined with detection network (e.g., Faster-RCNN) or segmentation network (i.e. CNN-based) for more generalized vision tasks in the wild. Besides, VTAL has also been used for cross-domain visual retrieval~\cite{HuangFX2020cvpr}, person re-identification~\cite{ZhaiYP2020cvpr}, person search~\cite{JingYa2020cvpr}, image restoration and dehazing~\cite{Yanbo2020cvpr,ShaoYJ2020cvpr}. It is hopeful that VTAL will play an increasingly important role in other vision tasks with more heuristic models and algorithms towards transferable and universal representations.

\section{Conclusion and Outlook}
Transfer adaptation learning is an energetic research field which aims to learn domain adaptive representations and classifiers from source domains toward representing and recognizing samples from a distribution different but semantic related target domain. This paper surveyed recent advances in transfer adaptation learning in the past decade and present a new taxonomy of five technical challenges being faced by researchers: instance re-weighting adaptation, feature adaptation, classifier adaptation, deep network adaptation and adversarial adaptation. Actually, a number of models are overlapped among the five subcategories, and the most appealing combination in most recent works is re-weighting+deep network+adversarial strategies, which are not specially discussed due to space limitation. Besides, 12 visual benchmarks that address multiple cross-domain recognition tasks are collected and summarized to help facilitate researchers' insight on the tasks and scenarios that transfer adaptation learning aims to address.

The proposed taxonomy of transfer adaptation learning challenges in this paper provides a framework for better understanding and identifying the status of the field, future research challenges and directions. Each challenge was summarized with a discussion of existing problems and future direction, which, we believe, are worth studying for better capturing general domain knowledge, toward universal machine learning. The typical problems of transfer learning and domain adaptation are \textit{negative transfer} (i.e., model over-fitting) and \textit{under-adaptation} (i.e., model under-fitting), which attracted many researchers' focus in the past decade, but not well addressed due to the difficulty in explicitly characterizing the domain discrepancy. There lacks of some judgment conditions and evaluation criteria. Negative transfer can be stated as \textit{transferring knowledge from source can have a negative impact on the target learner}. In~\cite{WangZ2019cvpr}, three impacting factors leading to underlying negative transfer are stressed, such as algorithmic choice, distribution divergence and the size of labeled target data. Throughout the entire research lines, one specific research area of transfer adaptation learning that seems to be still under-studied is the co-adaptation of multiple but heterogeneous domains, which goes beyond two homogeneous domains. This challenge is more approaching real-world scenarios that numerous domains can be found~\cite{Mancini2018CVPR}, and co-adaptation expects to capture \textit{commonality} and \textit{specificity} among multiple domains. Additionally, the essence of TAL is to learn domain-invariant but class-discriminative representation for \textit{unlabeled} target data regardless of close-set, open-set or partial DA. Therefore, unsupervised/self-supervised learning idea can be a choice used to improve TAL toward universal representations beyond the domain-invariant representations. To uncover the knowledge used to transfer, meta learning can be a choice towards explainable TAL. The meta-transfer learning in~\cite{JangY2019ICML} and task disentanglement~\cite{Zamir2018CVPR} show some instructive insights on learning what and where to transfer.

Undoubtedly, a huge and pleasant progress in TAL model and algorithm is witnessed with blowout publications, which has well answered \textit{how to facilitate domain adaptation under the well-established expected target error upper bound theory}. However, we are not optimistic about this and three open questions that remain to be unanswered are stressed. 1) \textit{When} will we need transfer adaptation learning for a given application scenario? The basic analyzing condition of whether cross-domain happens is still not clear, which makes TAL \textit{blind}. 2) \textit{Where} does the domain adaptation theory go when two domains are completely disjoint (i.e., large domain discrepancy) in open-world (dynamic) scenarios? The domain adaptation theory is scenario-dependent and the \textit{impossibility} theory~\cite{Ben-David2010Imposs} of TAL needs to be broken. Also, studying the target error \textit{lower-bound} may be more instructive than the expected error upper-bound, but seems to be paid less attention. 3) \textit{What} useful knowledge has been transferred from the source to target? Currently, it cannot be explicitly explained. We observe these heuristic but promising directions of transfer adaptation learning for future research. Addressing these issues and challenges can well improve the \textit{universality}, \textit{interpretability} and \textit{credibility} of TAL in \textit{open-world} scenarios toward \textit{safer} applications.
\section*{Acknowledgment}

The author would like to thank the pioneer researchers in transfer learning, domain adaptation and other related fields. The author would also like to thank Dr. Mingsheng Long and Dr. Lixin Duan for their kindly help in providing insightful discussions.

\ifCLASSOPTIONcaptionsoff
  \newpage
\fi



%

\bibliographystyle{IEEETran}
\bibliography{my1}

\begin{IEEEbiography}[{\includegraphics[width=1.2in,height=1.2in,clip,keepaspectratio]{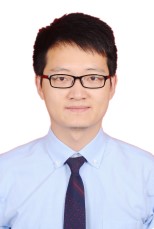}}]{Lei Zhang}
(M'14-SM'18) received his Ph.D degree in Circuits and Systems from the College of Communication Engineering, Chongqing University, Chongqing, China, in 2013. He was selected as a Hong Kong Scholar in China in 2013, and worked as a Post-Doctoral Fellow with The Hong Kong Polytechnic University, Hong Kong, from 2013 to 2015. He worked as a visiting professor at University of Macau in 2018. He is currently a Professor/Distinguished Research Fellow with Chongqing University. He has authored more than 100 scientific papers in top journals and conferences, including IEEE Transactions (e.g., T-PAMI, T-IP, T-MM, T-CSVT, T-NNLS), CVPR, ICCV, ECCV, ACM MM, AAAI, IJCAI, etc. He is on the Editorial Boards of several journals, such as IEEE Transactions on Instrumentation and Measurement, Neural Networks (Elsevier), etc. Dr. Zhang was a recipient of the 2019 ACM SIGAI Rising Star Award, the Best Paper Award of CCBR2017, Outstanding Doctoral Dissertation Award of Chongqing, China, in 2015, and the New Academic Researcher Award for Doctoral Candidates from the Ministry of Education, China, in 2012. His current research interests include deep learning, transfer learning, domain adaptation and computer vision. He is a Senior Member of IEEE.
\end{IEEEbiography}

\begin{IEEEbiography}[{\includegraphics[width=1in,height=1.25in,clip,keepaspectratio]{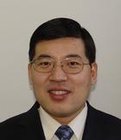}}]{Xinbo Gao}
received the B.Eng., M.Sc. and Ph.D. degrees in electronic engineering, signal and information processing from Xidian University, Xi'an, China, in 1994, 1997, and 1999, respectively. From 1997 to 1998, he was a research fellow at the Department of Computer Science, Shizuoka University, Shizuoka, Japan. From 2000 to 2001, he was a post-doctoral research fellow at the Department of Information Engineering, the Chinese University of Hong Kong, Hong Kong. Since 2001, he has been at the School of Electronic Engineering, Xidian University. He is currently a Cheung Kong Professor of Ministry of Education of P. R. China, a Professor of Pattern Recognition and Intelligent System of Xidian University and a Professor of Computer Science and Technology of Chongqing University of Posts and Telecommunications. His current research interests include Image processing, computer vision, multimedia analysis, machine learning and pattern recognition. He has published six books and around 300 technical articles in refereed journals and proceedings. Prof. Gao is on the Editorial Boards of several journals, including Signal Processing (Elsevier) and Neurocomputing (Elsevier). He served as the General Chair/Co-Chair, Program Committee Chair/Co-Chair, or PC Member for around 30 major international conferences. He is a Fellow of the Institute of Engineering and Technology and a Fellow of the Chinese Institute of Electronics.
\end{IEEEbiography}

\end{document}